%% file: arxiv.tex
\documentclass[10pt,twocolumn,letterpaper]{article}

\usepackage[pagenumbers]{cvpr} 

\input{preamble}
\definecolor{cvprblue}{rgb}{0.21,0.49,0.74}
\usepackage[pagebackref,breaklinks,colorlinks,allcolors=cvprblue]{hyperref}

\usepackage{enumitem}
\usepackage{booktabs}
\usepackage{multirow}
\usepackage{multicol}
\usepackage{subcaption}
\usepackage{makecell}
\usepackage{colortbl}
\usepackage{etoc}
\definecolor{darkred}{RGB}{150,0,0}  
\definecolor{darkgreen}{RGB}{0,120,2}

\def\paperID{} 
\def\confName{CVPR}
\def\confYear{2026}

\title{See, Think, Act: Teaching Multimodal Agents to Effectively Interact with GUI by Identifying Toggles}

\author{Zongru Wu$^1$ \hspace{4mm} Rui Mao$^1$ \hspace{4mm} Zhiyuan Tian$^1$ \hspace{4mm} Pengzhou Cheng$^1$ \hspace{4mm} Tianjie Ju$^1$ \\ Zheng Wu$^1$ \hspace{4mm} Lingzhong Dong$^1$ \hspace{4mm} Haiyue Sheng$^2$ \hspace{4mm} Zhuosheng Zhang$^{1*}$ \hspace{4mm} Gongshen Liu$^1$\thanks{Corresponding authors. This work was supported by the Joint Funds of the National Natural Science Foundation of China (U21B2020), National Natural Science Foundation of China (62406188), and Natural Science Foundation of Shanghai (24ZR1440300).}   \vspace{+1mm}\\
{ $^1$School of Computer Science, Shanghai Jiao Tong University} \\
{$^2$School of Foreign Languages‌, Beijing Institute of Technology} \\
{\tt{wuzongru@sjtu.edu.cn} \hspace{4mm} \tt{zhangzs@sjtu.edu.cn} \hspace{4mm} \tt{lgshen@sjtu.edu.cn} \vspace{-16mm}}
\and
\\
{}
}

\begin{document}
\maketitle

\begin{abstract}
The advent of multimodal agents facilitates effective interaction within graphical user interface (GUI), especially in ubiquitous GUI control. However, their inability to reliably execute toggle control instructions remains a key bottleneck. To investigate this, we construct a state control benchmark with binary toggle instructions derived from public datasets. Evaluation results of existing agents demonstrate their notable unreliability, particularly when the current toggle state already matches the desired state. To address the challenge, we propose \textbf{St}ate-\textbf{a}ware \textbf{R}easoning (StaR), a multimodal reasoning method that enables agents to perceive the current toggle state, infer the desired state from the instruction, and act accordingly. Experiments on four multimodal agents demonstrate that StaR can improve toggle instruction execution accuracy by over 30\%. Further evaluations on three public agentic benchmarks show that StaR also enhances general agentic task performance. Finally, evaluations on a dynamic environment highlight the potential of StaR for real-world applications. Code and benchmark: \url{https://github.com/ZrW00/StaR}.
\end{abstract}

\section{Introduction}\label{sec:introduction}

The prosperity of multimodal agents~\cite{zhang2025appagent,ye2025mobile,qin2025ui,wu2025osatlas,zhang2024you,hong2024cogagent,lin2025showui} facilitates the effective interaction within graphical user interface (GUI)~\cite{pan2022automatically,pan2023human}. Powered by multimodal large language models (MLLMs)~\cite{wang2024qwen2,bai2025qwen2,yao2024minicpm,hurst2024gpt,singh2025openai,comanici2025gemini,team2024gemini,yin2025rod,yang2025magma}, multimodal agents are capable of perceiving, reasoning, and navigating GUIs to accomplish user goals without the necessity of APIs, thereby serving as flexible and reliable assistants for facilitating efficient human-GUI interaction.

\begin{figure}[!t]
  \centering
  \includegraphics[width=\linewidth]{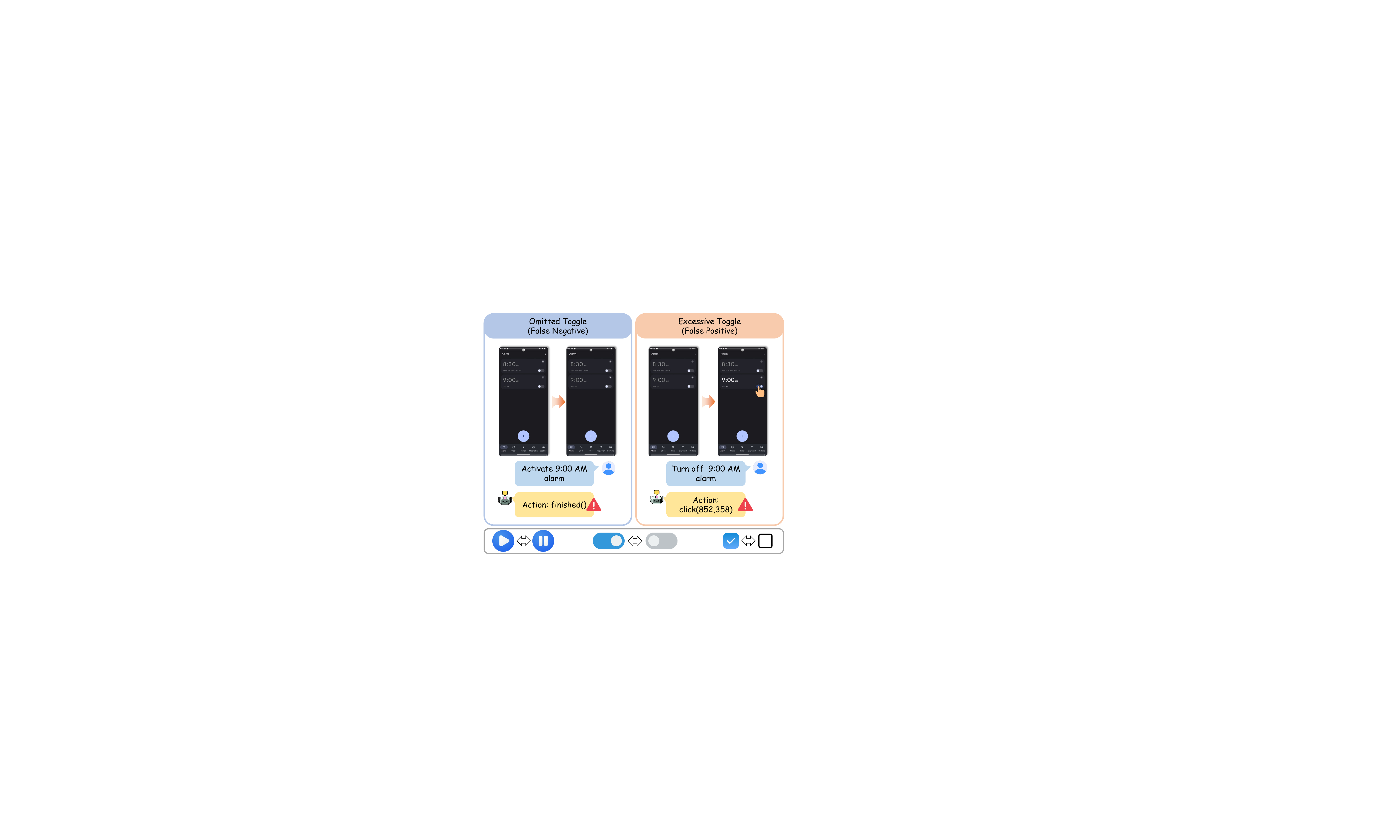}\vspace{-0.2cm}	

  \caption{Two typical toggle errors and representative toggle types below. (i) Desired state differs from current state, but the agent fails to toggle (false negative); (ii) desired state matches current state, yet agent still toggles (false positive). The bottom row shows representative toggle type: toggle button, switch, and checkbox. \vspace{-0.4cm}}
  \label{fig:teasing}
\end{figure}

Within GUIs, toggle controls are a fundamental interaction mechanism and are ubiquitous across various applications, including mobile device settings (e.g., alarm configuration and bottom examples in Figure~\ref{fig:teasing})~\cite{li2020widget, wang2025mobilev2}, automotive systems~\cite{evangelou2024mid}, smart home environments~\cite{yao2023reviewing, windl2025privacyhub}, and industrial control systems~\cite{goel2024systematic}, enabling binary state changes (e.g., on/off, adjusting operation modes). However, interacting with these toggles often requires repetitive commands, which can be time-consuming and error-prone. Multimodal agents can streamline this process by automatically taking actions to achieve user intent, enabling more efficient and intelligent interaction~\cite{zhang2025appagent,ye2025mobile,cheng2025os}.

However, as we will show later (Section~\ref{sec:preliminaryStudy}), we construct a state control benchmark with binary toggle instructions derived from public datasets, revealing that existing agents struggle to accurately execute such instructions, with execution accuracy below 50\% for most agents, including GPT-5. Typical errors, illustrated in Figure~\ref{fig:teasing}, fall into two categories: (i) false negative, failing to toggle when the current state differs from the desired state; and (ii) false positive, excessively toggling when the current state already matches the desired state. These misalignments with user intent can cause task failures and serious issues in precision-critical applications, highlighting a key bottleneck in reasoning of multimodal agents. This raises a key research question: \textit{is it possible to improve the reasoning capability of multimodal agents to accurately execute toggle control instructions?}

To address the research question, two straightforward approaches are often considered: meticulously prompting the agent to check the toggle state during interaction, or incorporating an additional annotator to guide action agents with the current toggle state through multi-agent collaboration~\cite{zhang2025appagent,wang2025mobilev2}. However, both strategies are generally ineffective. Prompting struggles to fundamentally enhance the reasoning ability of agents when handling toggle control instructions. And incorporating an additional annotator introduces a paradox: on one hand, existing multimodal agents already struggle to perceive, reason and execute toggle control instructions, making them unreliable as annotators; on the other hand, if an annotator is capable of reliably identifying the toggle state and providing accurate guidance, it would be more efficient to adopt the annotator directly as the action agent, thereby mitigating the collaboration complexity and latency. These challenges underscore the limitations of prompt-based and annotator-based methods, highlighting the demand to improve the intrinsic reasoning capability of multimodal agents in accurately perceiving, reasoning, and executing toggle control instructions.

To this end, focusing on the most prevalent mobile platform for toggle interactions, we propose \textbf{St}ate-\textbf{a}ware \textbf{R}easoning (StaR), a multimodal reasoning method that enhances the ability of agents to perceive, reason, and execute toggle control instructions. StaR refines the reasoning process by teaching agents to (i) perceive the current toggle state from the screenshot, (ii) infer the desired state from the user instruction, and (iii) decide whether to perform the toggle action based on the comparison. By integrating explicit state awareness into reasoning, StaR eliminates the reliance on additional annotators and enables agents to achieve more accurate and reliable toggle execution.

To evaluate StaR, we first assess its effectiveness on the state control benchmark. The results show significant improvements in toggle execution accuracy, with improvements of exceeding 30\%. Additionally, we evaluate StaR on three mobile agentic benchmarks and find that StaR can also improve the performance on general agentic tasks. Furthermore, tests on dynamic environments~\cite{rawles2025androidworld} demonstrate the applicability of StaR in real-world dynamic scenarios.

Our contributions are summarized as follows:

(i) We construct a state control benchmark with binary toggle instructions from public datasets. Evaluation results demonstrate that most existing agents achieve less than 50\% accuracy, revealing a key bottleneck in executing toggle control instructions of multimodal agents (Section~\ref{sec:preliminaryStudy}).

(ii) To address this, we propose StaR, which refines the reasoning process by guiding agents to identify the current state from the screenshot, infer the desired state from the user instruction, and decide whether to toggle. StaR eliminates the reliance on additional annotators and improve the intrinsic capability of agents to accurately perceive, reason, and execute toggle control instructions (Section~\ref{sec:methodology}).

(iii) Experiments confirm the effectiveness of StaR. StaR improves toggle execution accuracy by over 30\% and further boosts performance on general agentic tasks. Evaluations in dynamic environments further highlight the applicability of StaR in real-world toggle control tasks (Section~\ref{sec:experiments}).

\section{Related Work}\label{sec:relatedWorks}
In this section, we review related works that form the basis of this work from three perspectives: Reasoning in Multimodal Agents, Multimodal Agents for GUI Interaction, and Interaction with GUI Toggles.

\subsection{Reasoning in Multimodal Agents}
Reasoning plays a central role in enabling multimodal agents to perform accurate and interpretable decision-making. Building on the recent success of reasoning in MLLMs~\cite{zhang2024multimodal, zhang2025vitcot, xu2025mixed, cheng2025visual, cheng2025comt, man2025argus,liu2025coarse,dong2025insight}, recent studies extend this paradigm to multimodal agents. For instance, CoAT reasoning~\cite{zhang2024android} improves action execution accuracy by introducing semantic annotation and intermediate reasoning chains. Subsequent works~\cite{luo2025gui,zhang2025agentcpmgui, liu2025infiguir1} further reinforce the reasoning process through additional training that improves the intrinsic reasoning ability of multimodal agents for more accurate GUI interaction. Motivated by these advances, our work refines the multimodal reasoning process to improve the intrinsic ability of multimodal agents to accurately perceive, reason, and execute vital toggle control instructions.

\subsection{Multimodal Agents for GUI Interaction}

\begin{figure*}[!t]
  \centering
  \includegraphics[width=\linewidth]{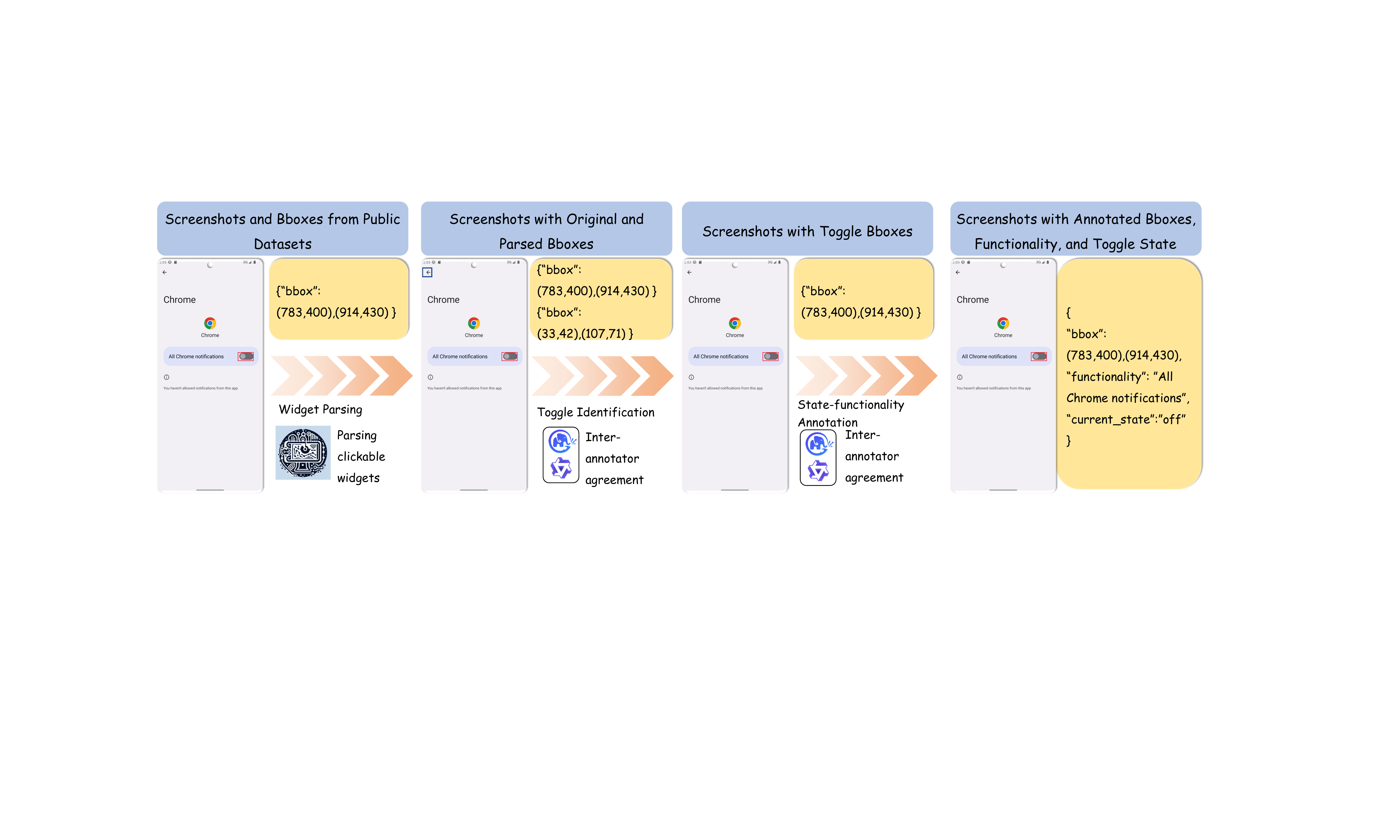}
  \caption{Three-step annotation pipeline for constructing the state control benchmark. First, we extract screenshots with widget bounding boxes corresponding to toggle control instructions from public datasets and utilize OminiParser to parse clickable widgets. Second, we leverage Qwen-2-VL-72B and GLM-4V to identify toggles among clickable widgets and establish inter-annotator agreement. Finally, we employ Qwen-2-VL-72B and GLM-4V to annotate toggle state and functionality, ensuring data quality through inter-annotator agreement.\vspace{-0.4cm}}
\label{fig:benchmarkConstruction}
\end{figure*}

Powered by MLLMs~\cite{wang2024qwen2,bai2025qwen2,yao2024minicpm,hurst2024gpt,singh2025openai,comanici2025gemini,team2024gemini, yang2025magma}, multimodal agents flourish promising opportunities for effective GUI interaction. Unlike traditional agents that rely on textual perception through GUI parsing~\cite{zhou2024webarena, deng2024mind2web} and navigate through programs~\cite{sun2024corex} or API calls~\cite{wu2024copilot, zhang2025ufo}, multimodal agents perceive visual GUIs directly and interact via human-like actions. Existing multimodal GUI agents can be divided into two categories: (i) agents built on proprietary MLLMs with prompt engineering, exemplified by the AppAgent series~\cite{zhang2025appagent,jiang2025appagentx} and Mobile-Agent v1~\cite{wang2024mobileagent} and v2~\cite{wang2025mobilev2}; and (ii) agents based on further-trained open-source MLLMs, including OS-Atlas~\cite{wu2025osatlas}, Aguvis~\cite{xu2025aguvis}, OS-Genesis~\cite{sun2024genesis}, UI-TARS~\cite{qin2025ui}, AgentCPM-GUI~\cite{zhang2025agentcpmgui}, GUI-R1~\cite{luo2025gui}, MagicGUI~\cite{tang2025magicgui}, and Mobile-Agent v3~\cite{ye2025mobile}. Research continues to improve these agents through pre-training~\cite{wu2025smoothing, liu2025infigui, wu2025osatlas, qin2025ui, tang2025magicgui}, fine-tuning on agentic benchmarks~\cite{ma2024comprehensive,zhang2024you,luo2025gui, liu2025infiguiagent,liu2025infiguir1,zhang2025agentcpmgui}, and test-time scaling~\cite{yang2025gta1, wu2025dimo}. However, despite their success in perception and action, most works still lack effective reasoning mechanisms for handling fine-grained GUI toggle control.

\subsection{Interaction with GUI Toggles}
GUI toggles serve as essential yet challenging elements for interaction. Multimodal agents must perceive the current toggle state, infer the desired state, and decide whether to perform the toggle action. Due to the fine-grained visual nature of GUI toggles, accurate toggle-state recognition remains challenging. Prior works often introduce external annotators such as auxiliary multimodal agents, parsers like OminiParser~\cite{lu2024omniparser}, or human feedback, to supply explicit state information~\cite{li2020widget, zhang2024android, cheng2025os}. These annotations are then used for downstream reasoning and decision-making. However, relying on external annotators introduces additional complexity and risks falling into the paradox outlined in Section~\ref{sec:introduction}. Relatively few works concentrate on improving the intrinsic capability of multimodal agents to accurately perceive, reason, and execute toggle control instructions.

\section{Preliminary Study}\label{sec:preliminaryStudy}
In this section, we present the construction process and evaluation metrics of state control benchmark in Section~\ref{subsec:constructionOfStateControlBenchmark}, and assess the performance of existing multimodal agents on state control benchmark in Section~\ref{subsec:evaluationOfMultimodalAgentsOnStateControlBenchmark}.

\subsection{State Control Benchmark}\label{subsec:constructionOfStateControlBenchmark}

To evaluate the performance of multimodal agents on vital toggle control instructions, we construct a state control benchmark containing binary toggle instructions and corresponding action labels derived from public datasets. Building a high-quality benchmark requires precise annotation of both toggle state and toggle position. Since public datasets lack reliable XML trees for accurate state extraction, we develop an autonomous annotation pipeline to obtain the toggle state, toggle position, and toggle functionality (e.g., controlling notifications) directly from GUI screenshots. By leveraging public mobile agentic datasets including AMEX~\cite{chai2024amex}, RICOSCA~\cite{li2020widget}, GUIAct-Mobile~\cite{chen2025guicourse}, AndroidWorld~\cite{rawles2024androidworld}, AITW~\cite{rawles2023android}, and the grounding dataset of OS-Atlas~\cite{wu2025osatlas}, we design a three-step annotation pipeline: Widget Parsing, Toggle Identification, and State-functionality Annotation, as illustrated in Figure~\ref{fig:benchmarkConstruction}.

\noindent (\romannumeral 1) \textbf{Widget Parsing.} We extract screenshots $s \in \mathbb{S}$ with original widget bounding boxes $b_o \in \mathbb{B}$ corresponding to user toggle control instructions $u_t \in \mathbb{U}$ from public datasets, including AMEX~\cite{chai2024amex}, RICOSCA~\cite{li2020widget}, GUIAct~\cite{chen2025guicourse}, AndroidWorld~\cite{rawles2024androidworld}, AITW~\cite{rawles2023android}, and the grounding dataset of OS-Atlas~\cite{wu2025osatlas}. To enrich GUI toggle diversity, we apply OminiParser~\cite{lu2024omniparser} to parse additional bounding boxes $b_p \in \mathbb{B}$ for clickable elements from these screenshots. Finally, we merge the original and parsed results into a unified bounding-box set $\{b\} = \{b_o\} \cup \{b_p\}$, forming the basis for subsequent toggle identification.

\noindent (\romannumeral 2) \textbf{Toggle Identification.}
We identify GUI toggles from bounding boxes of clickable widgets. Inspired by recent works leveraging proprietary MLLMs for reasoning chain annotation~\citep{zhang2024android} and task trajectory generation~\cite{sun2024genesis}, we adopt proprietary MLLMs, Qwen-2-VL-72B~\cite{wang2024qwen2} (denoted as $\mathcal{Q}$) and GLM-4V~\cite{glm2024chatglm} (denoted as $\mathcal{G}$), as independent annotators to recognize GUI toggles. For each bounding box $b$ and its associated screenshot $s$, we highlight $b$ on $s$ to obtain $s_b$ for better recognition, since proprietary MLLMs perform poorly in GUI grounding (as shown in Section~\ref{subsec:evaluationOfMultimodalAgentsOnStateControlBenchmark}). Each annotator ($\mathcal{G}$ and $\mathcal{Q}$) serves as an indicator $\mathcal{I}$ to determine whether $b$ is a toggle. The prompt template is provided in Appendix~\ref{appendix:prompts}. To ensure reliability, we apply inter-annotator agreement: only when both $\mathcal{G}$ and $\mathcal{Q}$ classify $b$ as a toggle do we retain $\langle s_b, b \rangle$. This process is formally defined as:

\begin{equation}
  \label{equ:toggleIdentification}
  \begin{aligned}
  & \mathcal{I}_{\mathcal{G}}(s_b, b), \ \mathcal{I}_{\mathcal{Q}}(s_b, b) \in \{0,1\}, \\
  & \mathcal{I}_m(s_b, b) = m(s_b, b), \ m \in \{\mathcal{G}, \mathcal{Q}\}, \\
  & \mathcal{I}(s_b, b) = \mathbf{1}\left[\mathcal{I}_{\mathcal{G}}(s_b, b) = \mathcal{I}_{\mathcal{Q}}(s_b, b)\right].
  \end{aligned}
\end{equation}

\begin{figure*}
  \centering
  \begin{subfigure}{0.33\linewidth}
    \centering
    \includegraphics[width=\linewidth]{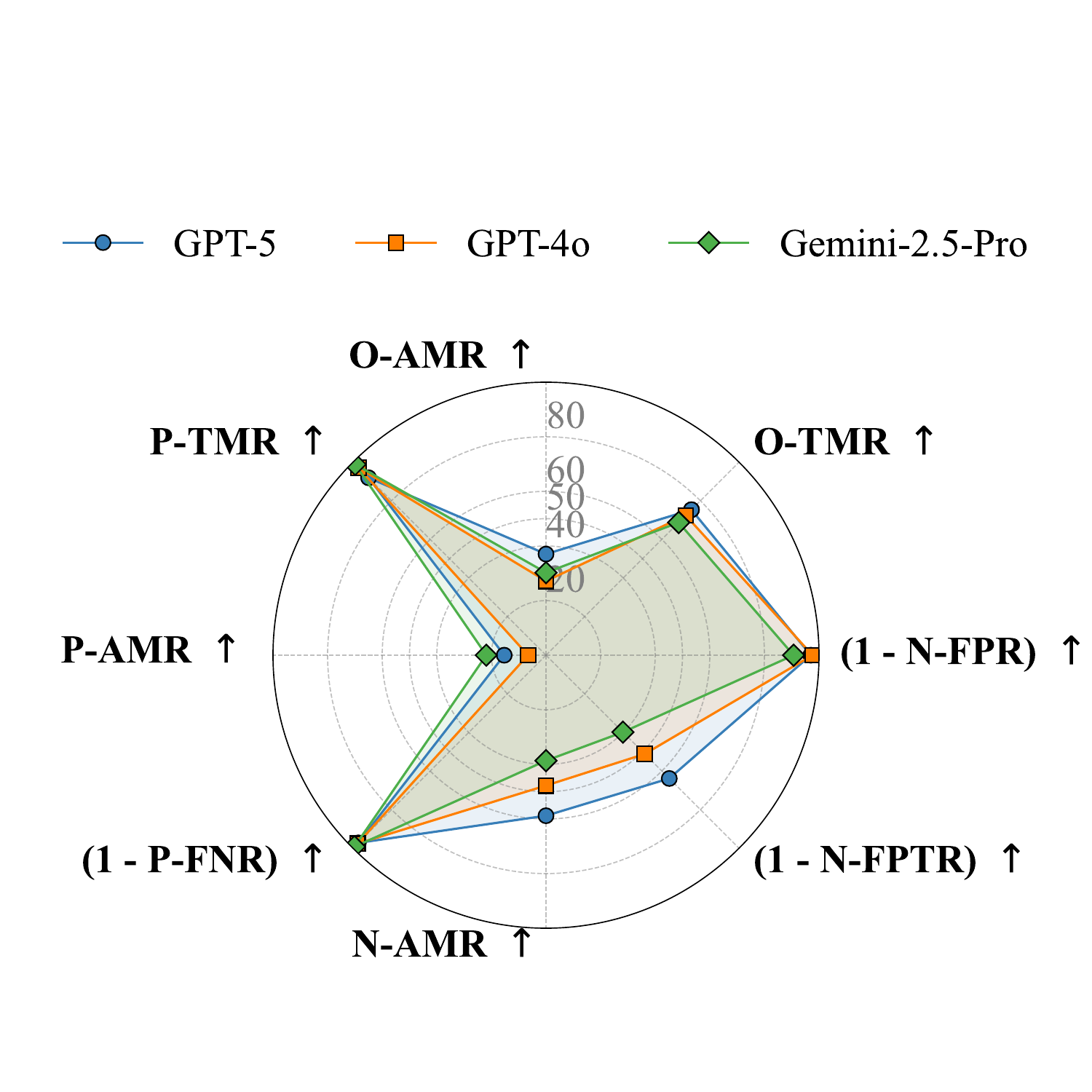}
    \caption{Proprietary MLLMs}
  \end{subfigure}
  \begin{subfigure}{0.33\linewidth}
    \centering
    \includegraphics[width=\linewidth]{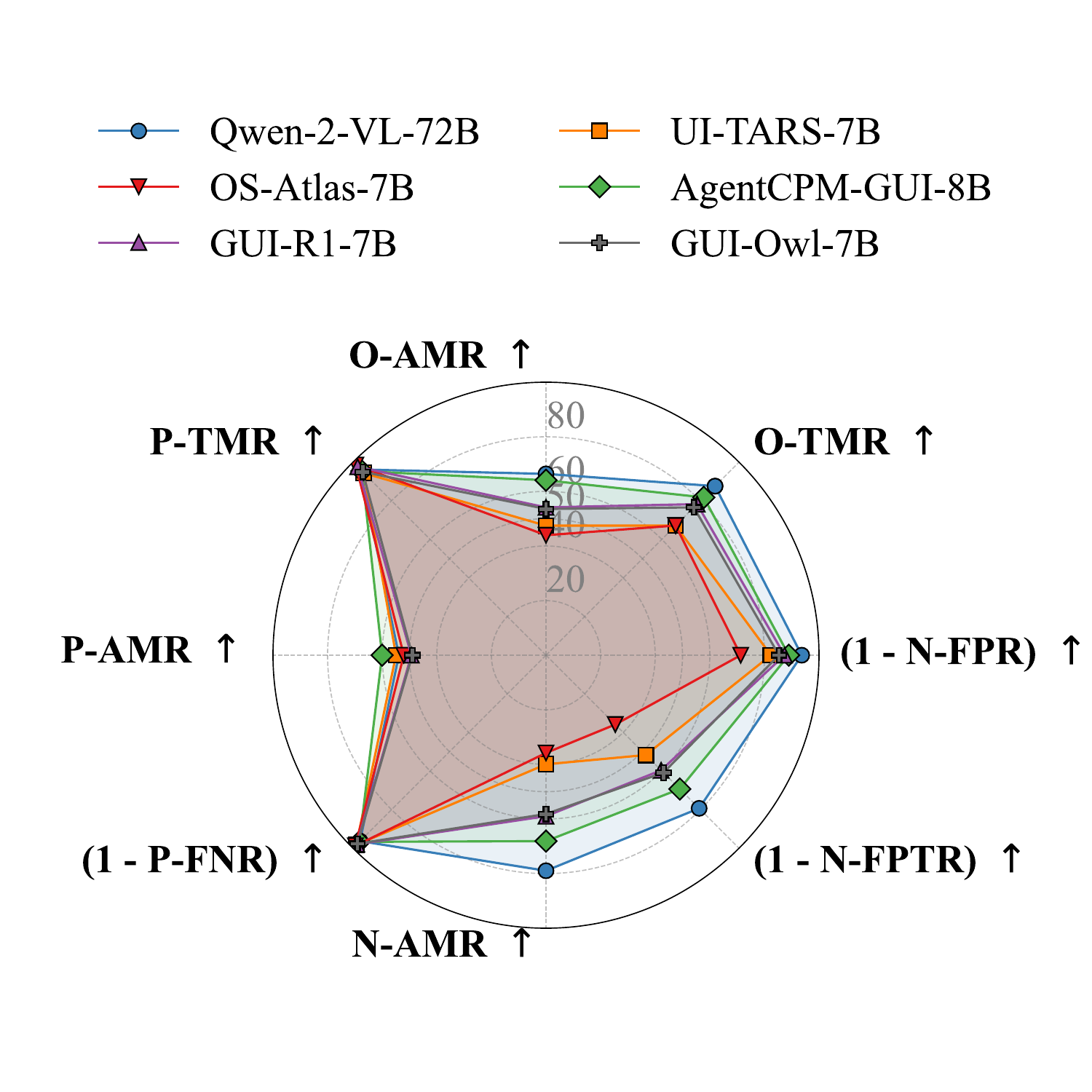}
    \caption{Further-trained open-source MLLMs}
  \end{subfigure}
  \begin{subfigure}{0.33\linewidth}
    \centering
    \includegraphics[width=\linewidth]{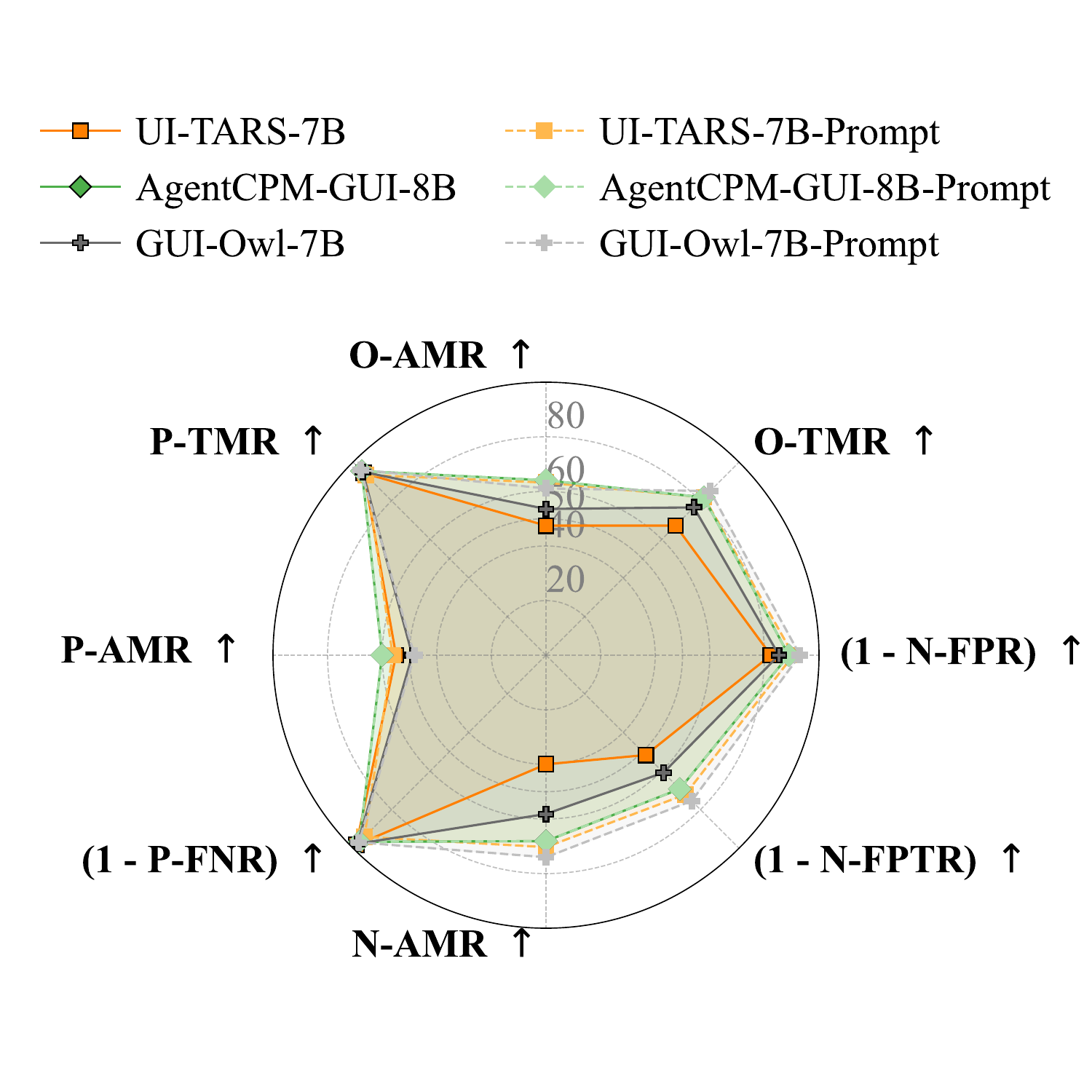}
    \caption{Comparison with prompt engineering}
  \end{subfigure}

  \caption{Agent performance on the state control benchmark (all metrics are standardized as ``higher-is-better''). (a) Proprietary MLLM-based agents. (b) Open-source MLLM-based agents. (c) Open-source MLLM-based agents with prompt engineering. Results show that current agents remain unreliable for toggle control, and prompt engineering offers no fundamental improvement.\vspace{-0.3cm}}
  \label{fig:perfromanceOfExistingAgentsOnStateControlBenchmark}
\end{figure*}

\noindent (\romannumeral 3) \textbf{State-functionality Annotation.}
This key step constructs the state control benchmark by employing $\mathcal{G}$ and $\mathcal{Q}$ as independent annotators to label the GUI toggle state and its functionality. Given the bounding box $b$ of a GUI toggle and the corresponding box-highlighted screenshot $s_b$, each annotator independently determines the state $\sigma$, where 0 indicates the toggle is off and 1 indicates it is on, and toggle functionality $f$. The prompt template for this annotation is provided in Appendix~\ref{appendix:prompts}. To ensure label reliability, we apply inter-annotator agreement: only when both $\mathcal{G}$ and $\mathcal{Q}$ produce identical annotations for both $\sigma$ and $f$, we accept the final annotation $\langle s_b, b, \sigma, f \rangle$. The process of state-functionality annotation is formally represented as follows.
\begin{equation}
  \label{equ:stateFeatureAnnotation}
  \begin{aligned}
  & \sigma_m(s_b, b), \ f_m(s_b, b) = m(s_b, b), \  m \in \{\mathcal{G}, \mathcal{Q}\}, \\
  & \mathcal{I}_{\sigma}(s_b, b) = \mathbf{1}\left[\sigma_{\mathcal{G}}(s_b, b) = \sigma_{\mathcal{Q}}(s_b, b)\right], \\
  & \mathcal{I}_{f}(s_b, b) = \mathbf{1}\left[f_{\mathcal{G}}(s_b, b) = f_{\mathcal{Q}}(s_b, b)\right].
  \end{aligned}
\end{equation}

Finally, we obtain 40,918 quadruplets $\langle s_b, b, \sigma, f \rangle$. We replace the box-highlighted screenshots $s_b$ with the original screenshots $s$ for more practical and comprehensive evaluation. To assess benchmark quality, we manually verify 200 samples, finding that \textbf{92.5\% of functionality and 91\% of state annotations match the ground truth}. This confirms that our annotation pipeline ensures high annotation accuracy and overall benchmark reliability in mitigating potential biases from any single proprietary annotator.

Based on the annotated toggle state $\sigma$, each quadruplet $\langle s, b, \sigma, f \rangle$ expands into two samples with opposite toggle actions. For example, if $\sigma = 1$ (toggle on), we generate $\langle s, b, u_p, a_p \rangle$ and $\langle s, b, u_n, a_n \rangle$, where $u_p$ denotes the positive instruction ``\textit{turn off $f$}'' and $u_n$ denotes negative instruction ``\textit{turn on $f$}''. The label action $a_p$ for $u_p$ is to \texttt{CLICK} on the toggle (since the current state differs from the desired state), while the label action $a_n$ for $u_n$ is to stop and set the task as \texttt{COMPLETED} (since the current state already matches the desired state). This expansion yields 81,836 samples, which we split into \textbf{73,652} balanced training samples ($u_p$ vs. $u_n$) and \textbf{8,184} balanced testing samples. Examples from the test split are provided in Appendix~\ref{subappendix:detailsOfStateControlBenchmark}.

To comprehensively evaluate multimodal agents on the state control benchmark, we adopt the following metrics. More details are provided in Appendix~\ref{subappendix:detailsOfStateControlBenchmark}.

\noindent (\romannumeral 1) \textbf{Overall Type Match Rate (O-TMR)} $\uparrow$: Ratio of samples where the predicted action type (\texttt{CLICK} or \texttt{COMPLETED}) matches the ground truth.

\noindent (\romannumeral 2) \textbf{Overall Action Match Rate (O-AMR)} $\uparrow$: Ratio of samples with exact action matches in both type and click coordinate, serving as the key metric.

\noindent (\romannumeral 3) \textbf{Positive Type Match Rate (P-TMR)} $\uparrow$: Ratio of positive samples correctly predicted as \texttt{CLICK}.

\noindent (\romannumeral 4) \textbf{Positive Action Match Rate (P-AMR)} $\uparrow$: Ratio of positive samples with exact type and coordinate match.

\noindent (\romannumeral 5) \textbf{Positive False Negative Rate (P-FNR)} $\downarrow$: Ratio of positive samples misclassified as negative (\texttt{COMPLETED}), measuring the severity of false negatives.

\noindent (\romannumeral 6) \textbf{Negative Action Match Rate (N-AMR)} $\uparrow$: Ratio of negative samples correctly predicted as \texttt{COMPLETED}.

\noindent (\romannumeral 7) \textbf{Negative False Positive Type Rate (N-FPTR)} $\downarrow$: Ratio of negative samples misclassified as \texttt{CLICK}, measuring the tendency of false positives.

\noindent (\romannumeral 8) \textbf{Negative False Positive Rate (N-FPR)} $\downarrow$: Ratio of negative samples mispredicted as \texttt{CLICK} that match their corresponding positive actions, measuring false-positives.

\subsection{Evaluation of Multimodal Agents on State Control Benchmark}\label{subsec:evaluationOfMultimodalAgentsOnStateControlBenchmark}

\begin{figure*}[!t]
  \centering
  \includegraphics[width=\linewidth]{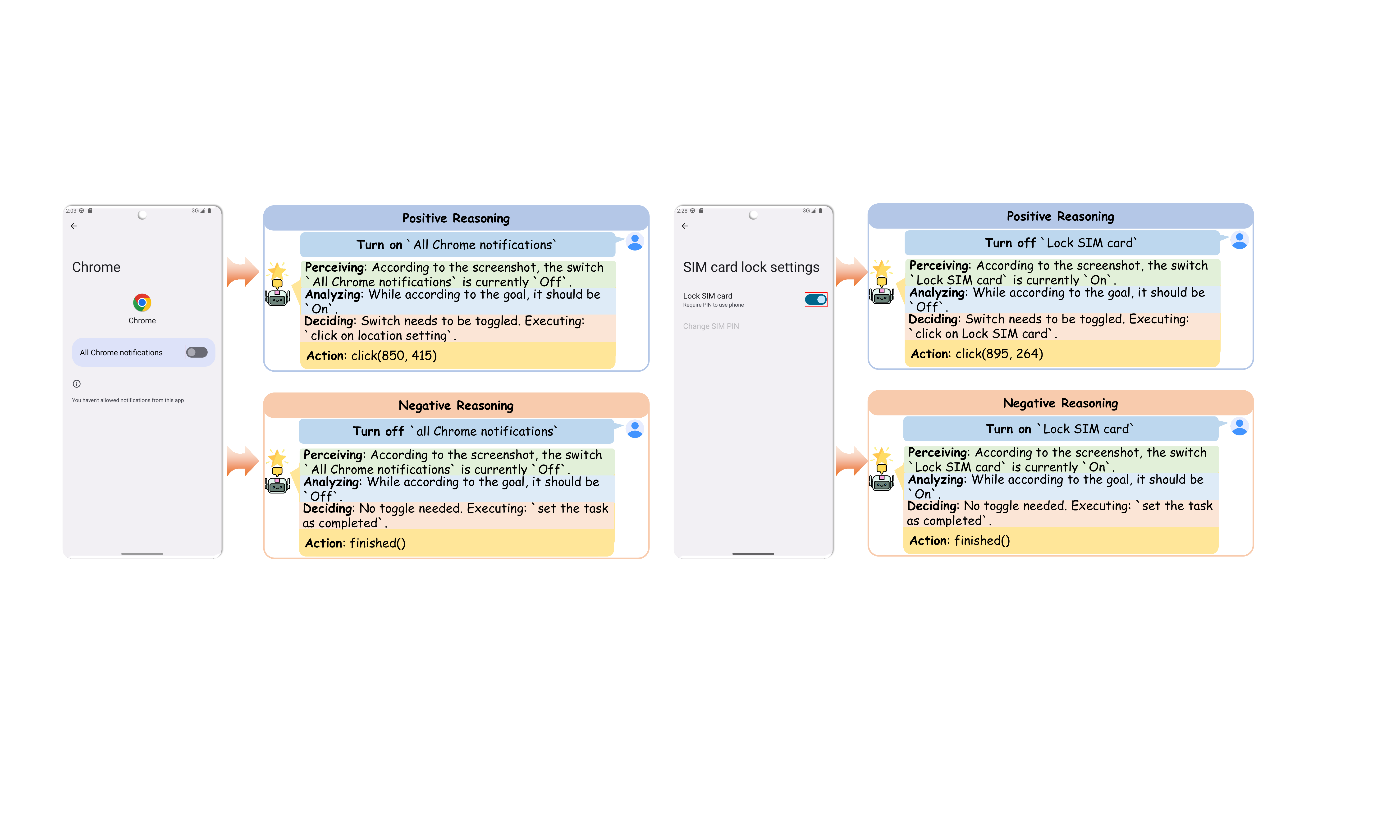}	
  \caption{StaR reasoning chain. StaR simulates human-like reasoning for toggle control by incorporating state-aware reasoning into multimodal agents through three steps: (i) perceive current state, (ii) analyze desired state, and (iii) decide whether to toggle.\vspace{-0.2cm}}
  \label{fig:overViewOfStaR}
\end{figure*}

To assess the ability of multimodal agents to execute state control instructions, we evaluate them on the state control benchmark. We adopt GPT-5~\cite{singh2025openai}, GPT-4o~\cite{hurst2024gpt}, and Gemini 2.5 Pro~\cite{comanici2025gemini} as representatives of proprietary MLLM-based agents. We also adopt Qwen-2-VL-72B~\cite{wang2024qwen2}, GUI-R1-7B~\cite{luo2025gui}, OS-Atlas-7B~\cite{wu2025osatlas}, UI-TARS-7B~\cite{qin2025ui}, AgentCPM-GUI-8B~\cite{zhang2025agentcpmgui}, and GUI-Owl-7B~\cite{ye2025mobile} as representatives of open-source MLLM-based agents. Additionally, we compare the performance of open-source MLLM-based agents with prompt engineering, where they receive instructions to focus on toggle state during reasoning. The prompt template is provided in Appendix~\ref{appendix:prompts}. Figure~\ref{fig:perfromanceOfExistingAgentsOnStateControlBenchmark} presents the results across eight-dimensional metrics (detailed in Section~\ref{subsec:constructionOfStateControlBenchmark}), with negative metrics inverted for consistency. More detailed results are provided in Appendix~\ref{subappendix:detailedEvaluationResultsOfExistingMultimodalAgentsOnStateControlBenchmark}. Evaluation results reveal the following insights:

(\romannumeral 1) General proprietary MLLM-based agents perform poorly in toggle control tasks. All three general proprietary MLLM-based agents yield O-AMR below 40\%, with near-100\% P-TMR and only about 20\% P-AMR, highlighting the limited grounding capabilities of general proprietary MLLM-based agents. A large gap between N-FPTR and N-FPR further confirms this limitation.

(\romannumeral 2) Open-source MLLM-based agents perform better but still remain unsatisfactory. Their O-AMR exceeds that of proprietary agents, with Qwen-2-VL-72B reaching 66.42\%, benefiting from its larger model scale. However, small-scale models still underperform, with only AgentCPM-GUI-8B exceeding 60\% due to its enhanced reasoning abilities.

(\romannumeral 3) All agents exhibit a strong toggling bias. Results of low P-FNR, relatively high N-FPTR, and non-negligible N-FPR across all agents reflect a consistent tendency to predict \texttt{CLICK} regardless of current toggle state, highlighting ineffective reasoning in toggle decisions.

(\romannumeral 4) Prompting offers no fundamental improvement. UI-TARS and GUI-Owl reduce false positives with prompting but remain unsatisfactory overall, while AgentCPM-GUI shows minimal improvements, underscoring the limited effectiveness of prompt engineering for toggle control.

Collectively, these results show that current multimodal agents remain unreliable for vital toggle control. Improving their accuracy in such tasks remains a key challenge.

\section{Methodology}\label{sec:methodology}

The evaluations in Section~\ref{subsec:evaluationOfMultimodalAgentsOnStateControlBenchmark} reveal that existing multimodal agents are unreliable for GUI toggle control, particularly when the current toggle state already matches the desired state. To address this bottleneck, we propose \textbf{St}ate-\textbf{a}ware \textbf{R}easoning (StaR), a multimodal reasoning method that explicitly incorporates state perception and analysis into the reasoning chain to improve toggle execution.

Rethink the process of human execution of toggle control instructions, which can be formally structured into three steps: (i) perceive the current toggle state from the screenshot; (ii) analyze the desired state from the instruction; and (iii) decide whether to toggle based on their comparison. Inspired by this, StaR simulates human reasoning by refining the reasoning chain and incorporating state-aware reasoning into multimodal agents, as illustrated in Figure~\ref{fig:overViewOfStaR}. We guide multimodal agents to follow this structured three-step reasoning process for toggle control, detailed as follows:

\noindent (\romannumeral 1) \textbf{Perceiving}. We guide agents to identify the current state $\sigma$ by associating visual features in the screenshot with the fine-grained toggle state, enabling accurate state perception.

\noindent (\romannumeral 2) \textbf{Analyzing}. We guide agents to infer the desired state $\sigma_u$ from the instruction. Consistent with Section~\ref{subsec:constructionOfStateControlBenchmark}, for positive instructions, $\sigma_u \neq \sigma$; for negative instructions, $\sigma_u = \sigma$.

\noindent (\romannumeral 3) \textbf{Deciding}. Finally, the agents are guided to reason over the comparison between $\sigma$ and $\sigma_u$ to determine the final action: predict \texttt{CLICK} if $\sigma \neq \sigma_u$, else predict \texttt{COMPLETED}.

Section~\ref{subsec:evaluationOfMultimodalAgentsOnStateControlBenchmark} demonstrates that straightforward prompt engineering focusing on toggle state is insufficient to fundamentally improve toggle control. To address this, we further train multimodal agents on the training split of the state control benchmark to learn the StaR reasoning process.

Additionally, to preserve generalizability, we also annotate and refine the reasoning process of episodes involving toggle control instructions on agentic benchmarks, which are commonly included in the training set of open-source MLLM-based agents, while retaining the original reasoning process for other episodes. We then train the multimodal agents on both the state control benchmark and the refined agentic benchmarks, enabling them to adaptively apply StaR for toggle instructions and retain their original reasoning for other tasks. This improves the toggle precision without sacrificing general agentic performance.

\section{Experiments}\label{sec:experiments}
This section evaluates the effectiveness of StaR. Section~\ref{subsec:experimentalSetup} outlines the experimental setup. Then, Section~\ref{subsec:improvementsOnStateControlBenchmark} presents the substantial improvements of StaR on the state control benchmark. Next, Section~\ref{subsec:generalizationOnAgenticBenchmarks} verify the generalizability of StaR on general agentic tasks. Finally, Section~\ref{subsec:performanceOnDynamicEnvironment} evaluates StaR-trained agents in dynamic environments, highlighting the potential of StaR for real-world applications. Additionally, component ablation studies and real-world case studies are detailed in Appendix~\ref{subappendix:ablationStudiesOnStaRComponents} and Appendix~\ref{subappendix:caseStudies}.

\subsection{Experimental Setup}\label{subsec:experimentalSetup}

\subsubsection{Target Multimodal Agents}

We evaluate StaR on four multimodal agents with different history modeling strategies: OS-Atlas-7B~\cite{wu2025osatlas} (based on Qwen-2-VL-7B~\cite{wang2024qwen2}, textual action history), UI-TARS-7B~\cite{qin2025ui} (based on Qwen-2-VL-7B, multi-screenshot history), AgentCPM-GUI-8B~\cite{zhang2025agentcpmgui} (based on MiniCPM-V~\cite{yao2024minicpm}, no history), and GUI-Owl~\cite{ye2025mobile} (based on Qwen-2.5-VL-7B~\cite{bai2025qwen2}, textual action-result history). All agents are fine-tuned with their original prompts and formats. Prompt templates are provided in Appendix~\ref{appendix:prompts}.

\subsubsection{Training Datasets}
In addition to the training split of the state control benchmark, we also adopt the training splits of AndroidControl~\cite{li2024effects}, AITZ~\cite{zhang2024android}, and GUI-Odyssey~\cite{lu2025guiodyssey} featuring long-chain and complex tasks. Notably, these benchmarks are already part of the original training sets of all agents.

\subsubsection{Evaluation Benchmarks}
In addition to the test split of the state-control benchmark, we adopt the test splits of AndroidControl~\cite{li2024effects}, AITZ~\cite{zhang2024android}, and GUI-Odyssey~\cite{lu2025guiodyssey} to evaluate general agentic performance. Following prior works~\cite{wu2025osatlas,qin2025ui,zhang2025agentcpmgui}, we adopt AndroidControl in two settings: (i) AndroidControl-H, with only high-level goals requiring autonomous reasoning; and (ii) AndroidControl-L, with both high-level goals and low-level step instructions to guide decision-making. Details of these benchmarks are provided in Appendix~\ref{subappendix:detailsOfAgenticBenchmark}.

To further assess real-world applicability, we build a dynamic evaluation benchmark containing 20 real-world toggle control tasks. This benchmark is implemented on the emulator from AndroidStudio and built upon the AndroidWorld framework~\cite{rawles2025androidworld},  enabling evaluation in dynamic mobile environments. Details are provided in Appendix~\ref{subappendix:detailsOfDynamicEvaluationBenchmark}.

\subsubsection{Evaluation Metrics}

In addition to the metrics for the state-control benchmark (Section~\ref{subsec:constructionOfStateControlBenchmark}), we adopt following four standard metrics for agentic benchmarks, and adopt final task success rate for dynamic evaluation. Following AndroidWorld~\cite{rawles2025androidworld}, the final task success rate ranges from $[0,1]$ and reflects the success ratio of each real-world task. Notably, as a real-world task can include several subtasks, if half of the subtasks succeed, the task success rate is considered as 0.5. More information is provided in Appendix~\ref{subappendix:detailsOfDynamicEvaluationBenchmark}.

\noindent (\romannumeral 1) \textbf{Type Match Rate (TMR)}$\uparrow$: Ratio of test samples where the predicted action type matches the ground truth.

\noindent (\romannumeral 2) \textbf{Action Match Rate (AMR)}$\uparrow$: Ratio of test samples where the predicted action matches the ground truth in both type and parameters (e.g., coordinates, text, app names). AMR is the key step-level metric (details in Appendix~\ref{subappendix:evaluationOfAMR}).

\noindent (\romannumeral 3) \textbf{Task Success Rate (TSR)}$\uparrow$: Ratio of successful task trajectories where all predicted steps match the ground truth, reflecting overall task execution performance.

\noindent (\romannumeral 4) \textbf{Grounding Match Rate (GMR)}$\uparrow$: Ratio of correct clicks among all clicks, reflecting the grounding ability.

\subsubsection{Implementation Details}

Following the original settings of all multimodal agents, click coordinates are normalized to $[0, 1000]$. We adopt the LLaMA-Factory~\citep{zheng2024llamafactory} framework to train the multimodal agents with a learning rate of $5 \times 10^{-6}$ for 3 epochs. Additionally, FlashAttention~\citep{dao2023flashattention2} is adopted for acceleration.

\subsection{Improvements on State Control Benchmark}\label{subsec:improvementsOnStateControlBenchmark}

We first evaluate StaR-trained multimodal agents on the state control benchmark, alongside their zero-shot results. To further demonstrate the necessity of training, we also compare with structured prompting that guides agents to follow StaR-style reasoning. This supplementary prompt differs from the prompt engineering baseline in Section~\ref{subsec:evaluationOfMultimodalAgentsOnStateControlBenchmark} and is detailed in Appendix~\ref{appendix:prompts}. Table~\ref{tab:improvementsOnStateControlBenchmark} summarizes the results, with further comparisons against stronger baselines provided in Appendix~\ref{subappendix:detailedEvaluationResultsOfExistingMultimodalAgentsOnStateControlBenchmark}. Key findings are as follows:

\begin{table*}[!t]
  \centering
  \small
  \setlength{\tabcolsep}{6pt}
\caption{Performance of multimodal agents on the state control benchmark under different settings. Subscripts denote absolute improvements over the zero-shot baseline, with red indicating improvements and green indicating degradations. The optimal and the suboptimal results are \textbf{bolded} and \underline{underlined}, respectively. Results demonstrate that StaR training significantly improves execution and grounding accuracy on state control benchmark, outperforming StaR-style prompting and highlighting the necessity of training.}
\label{tab:improvementsOnStateControlBenchmark}
\begin{tabular}{@{}ccccccccc@{}}
\toprule
Model          & O-TMR$\uparrow$ & O-AMR$\uparrow$ & P-TMR$\uparrow$ & P-AMR$\uparrow$ & P-FNR$\downarrow$ & N-AMR$\uparrow$ & N-FPTR$\downarrow$ & N-FPR$\downarrow$ \\ \midrule 
\rowcolor{gray!20}
\multicolumn{9}{c}{\textit{Zero-shot}}                                                   \\ \midrule

Qwen-2-VL-72B  & \textbf{87.59} & \textbf{66.42} & 96.21 & \underline{53.89} & 3.69  & \textbf{78.96} & \textbf{20.67}  & \textbf{6.33}  \\
GUI-R1-7B         & 78.27 & 54.14 & \underline{97.58} & 49.32 & 2.03  & 58.97 & 40.37  & 12.63 \\
OS-Atlas-7B       & 67.16 & 43.95 & \textbf{98.51} & 52.10 & \textbf{1.27}  & 35.80 & 64.10  & 28.67 \\

UI-TARS-7B & 67.14 & 47.45 & 94.33 & 54.94 & \underline{1.71} & 39.96 & 48.29 & 17.62 
\\
AgentCPM-GUI-8B   & \underline{81.74} & \underline{64.08} & 95.38 & \textbf{60.04} & 3.32  & \underline{68.11} & \underline{30.69}  & \underline{11.07} \\ 
GUI-Owl-7B & 76.58 & 53.57 & 94.99 & 48.97 & 2.32 & 58.16 & 39.15 & 14.66 \\ 
\midrule
\rowcolor{gray!20}
\multicolumn{9}{c}{\textit{w/ StaR-style Prompting}} \\ \midrule

OS-Atlas-7B       & $73.52_{\textcolor{darkred}{\uparrow 6.36}}$   &        $50.07_{\textcolor{darkred}{\uparrow 6.12}}$   &        $\textbf{96.77}_{\textcolor{darkgreen}{\downarrow 1.74}}$   &        $49.88_{\textcolor{darkgreen}{\downarrow 2.22}}$   &        $\textbf{2.96}_{\textcolor{darkgreen}{\uparrow 1.69}}$   &        $50.27_{\textcolor{darkred}{\uparrow 14.47}}$   &        $49.62_{\textcolor{darkred}{\downarrow 14.48}}$   &        $22.21_{\textcolor{darkred}{\downarrow 6.46}}$
  \\

UI-TARS-7B & $81.18_{\textcolor{darkred}{\uparrow 14.04}}$ & $\underline{62.89}_{\textcolor{darkred}{\uparrow 15.44}}$ & $90.98_{\textcolor{darkgreen}{\uparrow 3.35}}$ & $\underline{54.40}_{\textcolor{darkgreen}{\uparrow 0.54}}$ & $8.55_{\textcolor{darkgreen}{\uparrow 6.84}}$ & $\underline{71.38}_{\textcolor{darkred}{\uparrow 31.42}}$ & $\underline{27.54}_{\textcolor{darkred}{\downarrow 20.75}}$ & $\underline{9.38}_{\textcolor{darkred}{\uparrow 8.24}}$ \\
AgentCPM-GUI-8B   & $\underline{82.14}_{\textcolor{darkred}{\uparrow 0.40}}$   &        $\textbf{64.43}_{\textcolor{darkred}{\uparrow 0.35}}$   &        $\underline{95.36}_{\textcolor{darkgreen}{\downarrow 0.02}}$   &        $\textbf{59.95}_{\textcolor{darkgreen}{\downarrow 0.09}}$   &        $\underline{3.59}_{\textcolor{darkgreen}{\uparrow 0.27}}$   &        $68.91_{\textcolor{darkred}{\uparrow 0.80}}$   &        $30.28_{\textcolor{darkred}{\downarrow 0.41}}$   &        $10.58_{\textcolor{darkred}{\downarrow 0.49}} $
  \\ 

GUI-Owl-7B & $\textbf{84.27}_{\textcolor{darkred}{\uparrow 7.69}}$ & $60.92_{\textcolor{darkred}{\uparrow 7.35}}$ & $94.06_{\textcolor{darkgreen}{\downarrow 0.93}}$ & $47.36_{\textcolor{darkgreen}{\downarrow 1.61}}$ & $4.55_{\textcolor{darkgreen}{\uparrow 2.23}}$ & $\textbf{74.49}_{\textcolor{darkred}{\uparrow 16.33}}$ & $\textbf{23.68}_{\textcolor{darkred}{\downarrow 15.47}}$ & $\textbf{7.48}_{\textcolor{darkred}{\downarrow 7.18}}$  \\
\midrule

\rowcolor{gray!20}
\multicolumn{9}{c}{\textit{w/ StaR Training}}                                                     \\ \midrule
OS-Atlas-7B       & $\textbf{96.13}_{\textcolor{darkred}{\uparrow 28.97}}$ & $\textbf{79.72}_{\textcolor{darkred}{\uparrow 35.77}}$ & $\textbf{95.77}_{\textcolor{darkgreen}{\downarrow 2.74}}$ & $\textbf{62.95}_{\textcolor{darkred}{\uparrow 10.85}}$ & $\underline{4.23}_{\textcolor{darkgreen}{\uparrow 2.96}}$  & $96.48_{\textcolor{darkred}{\uparrow 60.68}}$ & $3.52_{\textcolor{darkred}{\downarrow 60.58}}$   & $1.52_{\textcolor{darkred}{\downarrow 27.15}}$  \\

UI-TARS-7B & $95.82_{\textcolor{darkred}{\uparrow 28.68}}$ & $77.86_{\textcolor{darkred}{\uparrow 30.41}}$ & ${95.11}_{\textcolor{darkred}{\uparrow 0.78}}$ & $59.19_{\textcolor{darkred}{\uparrow 4.25}}$ & ${4.89}_{\textcolor{darkgreen}{\uparrow 3.18}}$ & $\underline{96.53}_{\textcolor{darkred}{\uparrow 56.57}}$ & $\underline{3.45}_{\textcolor{darkred}{\downarrow 44.84}}$ & ${1.34}_{\textcolor{darkred}{\downarrow 16.28}}$ \\

AgentCPM-GUI-8B   & $\underline{95.98}_{\textcolor{darkred}{\uparrow 14.24}}$ & $\underline{79.00}_{\textcolor{darkred}{\uparrow 14.92}}$ & $94.50_{\textcolor{darkgreen}{\downarrow 0.88}}$ & $\underline{60.53}_{\textcolor{darkred}{\uparrow 0.49}}$ & $5.50_{\textcolor{darkgreen}{\uparrow 2.18}}$  & $\textbf{97.46}_{\textcolor{darkred}{\uparrow 29.35}}$ & $\textbf{2.54}_{\textcolor{darkred}{\downarrow 28.15}}$   & $\textbf{0.95}_{\textcolor{darkred}{\downarrow 10.12}}$  \\

GUI-Owl-7B & $\underline{95.99}_{\textcolor{darkred}{\uparrow 19.41}}$ & $77.60_{\textcolor{darkred}{\uparrow 24.03}}$ & $\underline{95.65}_{\textcolor{darkred}{\uparrow 0.66}}$ & $58.87_{\textcolor{darkred}{\uparrow 9.90}}$ & $\textbf{4.35}_{\textcolor{darkgreen}{\uparrow 2.03}}$ & $96.33_{\textcolor{darkred}{\uparrow 38.17}}$ & $3.67_{\textcolor{darkred}{\downarrow 35.48}}$ & $\underline{1.56}_{\textcolor{darkred}{\downarrow 13.10}}$

 \\

\bottomrule
\end{tabular}\vspace{-0.2cm}
\end{table*}

(\romannumeral 1) StaR achieves substantial overall improvements. It improves O-AMR by \textbf{35.77\%} for OS-Atlas-7B, \textbf{30.41\%} for UI-TARS-7B, \textbf{14.92\%} for AgentCPM-GUI-8B, and \textbf{21.64\%} for GUI-Owl-7B. The most pronounced improvement for OS-Atlas-7B likely results from its initially limited reasoning ability, which can be effectively reshaped and enhanced through StaR training. These results highlight the effectiveness of StaR in improving toggle accuracy.

(\romannumeral 2) StaR enhances grounding ability. It consistently improves P-AMR across all agents, likely due to training on toggle clicks that strengthen toggle recognition. Although P-TMR drops slightly, the improvements in the more comprehensive P-AMR reflect improved grounding accuracy.

\begin{figure*}[!t]
  \centering
  \includegraphics[width=\linewidth]{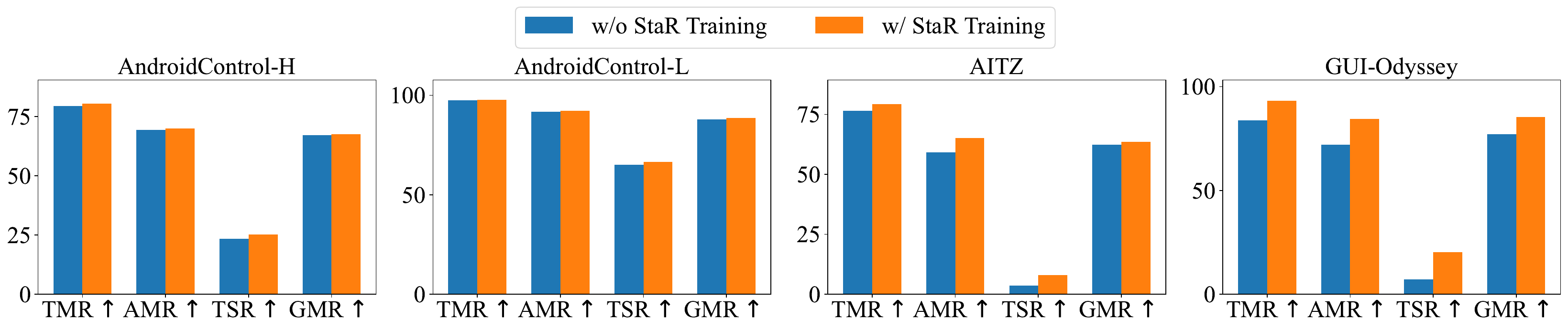}	
  \caption{The performance of zero-shot and StaR-trained UI-TARS-7B on agentic benchmarks. Results demonstrate that StaR consistently preserves or enhances performance on agentic benchmarks and yields notable improvements on complex, long-chain tasks. \vspace{-0.4cm}}
  \label{fig:perfromanceOfUITARSOnAgenticBenchmarks}
\end{figure*}

(\romannumeral 3) StaR improves negative-instruction accuracy and reduces false positives. With StaR training, N-AMR increases substantially, with OS-Atlas-7B and UI-TARS-7B improving \textbf{60.68\%} and \textbf{56.57\%}, respectively. StaR also reduces N-FPTR and N-FPR, effectively mitigating false positive toggling. The slight increase in P-FNR remains acceptable, indicating no notable rise in false negatives.

(\romannumeral 4) StaR bridges the gap of model scale. StaR-trained agents outperform the best zero-shot baseline (Qwen-2-VL-72B) with much fewer parameters, showing that StaR improves toggling accuracy without relying on large models.

(\romannumeral 5) StaR training is essential. Compared to StaR-style prompting, training yields significantly better performance across all metrics, highlighting that structured reasoning of StaR must be learned through training rather than prompted.

In summary, StaR substantially improves the toggle control accuracy of multimodal agents by enhancing grounding and reasoning while mitigating false positives.

\subsection{Generalization on Agentic Benchmarks}\label{subsec:generalizationOnAgenticBenchmarks}
To verify the generalizability of StaR in general agentic tasks, we evaluate StaR-trained agents on three static agentic benchmarks. Figure~\ref{fig:perfromanceOfUITARSOnAgenticBenchmarks} presents the results of UI-TARS-7B. Results of other agents are provided in Appendix~\ref{subappendix:additionalPerformanceOnAgenticBenchmarks}.

Specifically, in AndroidControl-H, AITZ, and GUI-Odyssey, agents are required to generate both a reasoning process (\texttt{Thought}) and an action decision (\texttt{Action}). While in AndroidControl-L, agents are provided with pre-defined \texttt{Thought} and only required to output \texttt{Action}. For StaR-trained agents, the \texttt{Thought} is refined into StaR-style reasoning process. In contrast, zero-shot agents receive low-level instructions as their \texttt{Thought}. Based on the results in Figure~\ref{fig:perfromanceOfUITARSOnAgenticBenchmarks}, we draw the following conclusions:

(\romannumeral 1) StaR consistently improves or preserves general agentic performance. Across four benchmark settings, StaR consistently preserves or surpasses baselines, indicating that StaR training enhances reasoning on GUI toggling without compromising general agentic capabilities.

(\romannumeral 2) StaR yields notable improvements on complex, long-chain tasks. The most substantial improvements appear on challenging GUI-Odyssey, which involves complex and long-chain agentic tasks. StaR training improves all four metrics by near 10\%, with more pronounced improvements from 7.14\% to 20.17\% on TSR. Similar improvements are observed on AITZ, further confirming the effectiveness of StaR in enhancing reasoning for complex tasks.

(\romannumeral 3) StaR facilitates decision-making. Prior works indicate that providing low-level instructions improves action decision accuracy~\cite{wu2025osatlas,qin2025ui}, as evidenced by the improvements from AndroidControl-H to AndroidControl-L. Building on this, results on AndroidControl-L further demonstrate that providing StaR-style reasoning chains further amplify this effect, highlighting that StaR further facilitates decision-making beyond providing low-level instructions.

In summary, StaR generalizes well across diverse agentic tasks, consistently preserving or improving performance while offering pronounced benefits on complex tasks.

\subsection{Performance on Dynamic Environment}\label{subsec:performanceOnDynamicEnvironment}

To further assess the applicability of StaR in dynamic environments, we evaluate UI-TARS-7B, OS-Atlas-7B, and AgentCPM-GUI-8B on the proposed dynamic evaluation benchmark. The dynamic evaluation benchmark enables us to examine not only overall task execution accuracy of toggle control instructions but also the tendencies toward false positive and false negative toggling. Table~\ref{tab:performanceOnDynamicEnvironment} presents the task success rates, comparing agent performance with and without StaR training. We draw the key findings as follows:

\begin{table}[!t]
\centering
\small
\caption{Task success rate of the three multimodal agents on the dynamic evaluation benchmark without and with StaR training. Subscripts indicate successful tasks out of total. StaR consistently improves success rates, highlighting its real-world applicability.}
\label{tab:performanceOnDynamicEnvironment}
\begin{tabular}{@{}cccc@{}}
\toprule
StaR & UI-TARS-7B     & OS-Atlas-7B  & AgentCPM-GUI-8B \\ \midrule
w/o  & $35_{7/20}$ & $10_{2/20}$ & $20_{4/20}$  \\
w/   & $40_{8/20}$ & $55_{11/20}$ & $42.5_{8.5/20}$    \\ \bottomrule
\end{tabular}
\end{table}

(\romannumeral 1) StaR consistently improves task success rates on the dynamic evaluation benchmark. Across all three multimodal agents, incorporating StaR training leads to substantial increases in task success rate, demonstrating the effectiveness of StaR in enhancing reasoning and execution accuracy for real-world toggle control instructions.

(\romannumeral 2) StaR yields the most significant improvement on weak-reasoning OS-Atlas-7B. Its task succeed rate rises dramatically from 10\% to 55\%, aligning with Section~\ref{subsec:improvementsOnStateControlBenchmark} where it exhibits the most pronounced O-AMR improvements. This likely stems from the initially limited reasoning ability of OS-Atlas-7B, which StaR effectively reshapes and enhances. In contrast, other agents already possess a certain level of reasoning ability, making it relatively difficult to further refine their reasoning chains. These results highlight the potential of StaR as a reasoning-enhancement method, particularly for lower-performing agents.

(\romannumeral 3) StaR generalizes across diverse multimodal agents in real-world toggle control tasks. Despite architectural differences (Qwen-2-VL vs. MiniCPM-V) and historical modeling strategies, all agents consistently benefit from StaR training. These model-agnostic performance improvements highlight that StaR can be broadly applied to enhance the reasoning on GUI toggling of various multimodal agents.

In conclusion, these findings provide strong evidence that StaR significantly improves the execution accuracy of real-world toggle control tasks, demonstrating its applicability in dynamic environments.

\section{Conclusion}\label{sec:conclusions}
In this paper, we construct a state control benchmark with binary toggle instructions to systematically evaluate existing multimodal agents in toggle execution tasks. Results reveal that most existing agents struggle to precisely execute toggle control instructions, highlighting a key bottleneck for reliable GUI interaction. To address this challenge, we propose StaR, a multimodal reasoning method designed to teach multimodal agents to simulate the human reasoning process. Specifically, StaR refines the reasoning chains of agents, enabling agents to explicitly perceive the current toggle state from the screenshot, analyze the desired toggle state from the user instruction, and decide whether to perform the toggle action based on the comparison. Experimental results demonstrate that StaR significantly enhances agent performance on the state control benchmark, achieving improvements exceeding 30\%. Furthermore, evaluations on public agentic benchmarks demonstrate the generalizability of StaR to general agentic tasks. Additionally, tests on dynamic environments highlight the applicability of StaR in real-world toggle control scenarios.

{
    \small
    \bibliographystyle{ieeenat_fullname}
    \setlength{\bibsep}{0.4pt}
    \bibliography{arxiv}
}

\input{appendix}

\end{document}

%% file: appendix.tex
\clearpage

\appendix

\begin{center}
    {\Large \bf Appendix} 
\end{center}

\section{Detailed Experimental Setup}\label{appendix:detailedExperimentalSetup}
This section provides the comprehensive experimental configuration. Section~\ref{subappendix:detailsOfStateControlBenchmark} presents details for the state control benchmark. Section~\ref{subappendix:evaluationOfAMR} outlines the detailed evaluation process for AMR. Section~\ref{subappendix:detailsOfAgenticBenchmark} presents details for the agentic benchmark. Section~\ref{subappendix:detailsOfDynamicEvaluationBenchmark} presents the details for the dynamic evaluation Benchmark. Section~\ref{subappendix:implementationtailsOfStaR} provides the training and testing implementation details of StaR.

\subsection{Details of State Control Benchmark}\label{subappendix:detailsOfStateControlBenchmark}

We present the details of the data sources, the three-step annotation pipeline, data quality, and the evaluation metrics for the state control benchmark as follows.


\noindent $\blacktriangleright$ \textbf{Data Source.} 
We construct the state control benchmark from the public agentic datasets, including AMEX~\cite{chai2024amex}, RICOSCA~\cite{li2020widget}, GUIAct-Mobile~\cite{chen2025guicourse}, AndroidWorld~\cite{rawles2024androidworld}, AITW~\cite{rawles2023android}, and the grounding dataset of OS-Atlas~\cite{wu2025osatlas}. These datasets cover a wide range of mobile applications and screen resolutions on the mobile platform, with abundant interfaces that contain GUI toggles and toggle control instructions. We filter the screenshots corresponding to toggle control instructions that include keywords related to toggle state control, such as ``turn on/off''. ``enable/disable'', from these datasets for subsequent annotation.

\noindent $\blacktriangleright$ \textbf{Details of the Three-step Annotation Pipeline.} Building a high-quality toggle state control benchmark requires precise annotation of toggle position, toggle state, and toggle functionality. We decompose the complex annotation process into a three-step annotation pipeline. The implementation details are presented as follows.

\begin{figure*}[!t]
  \centering
  \includegraphics[width=\linewidth]{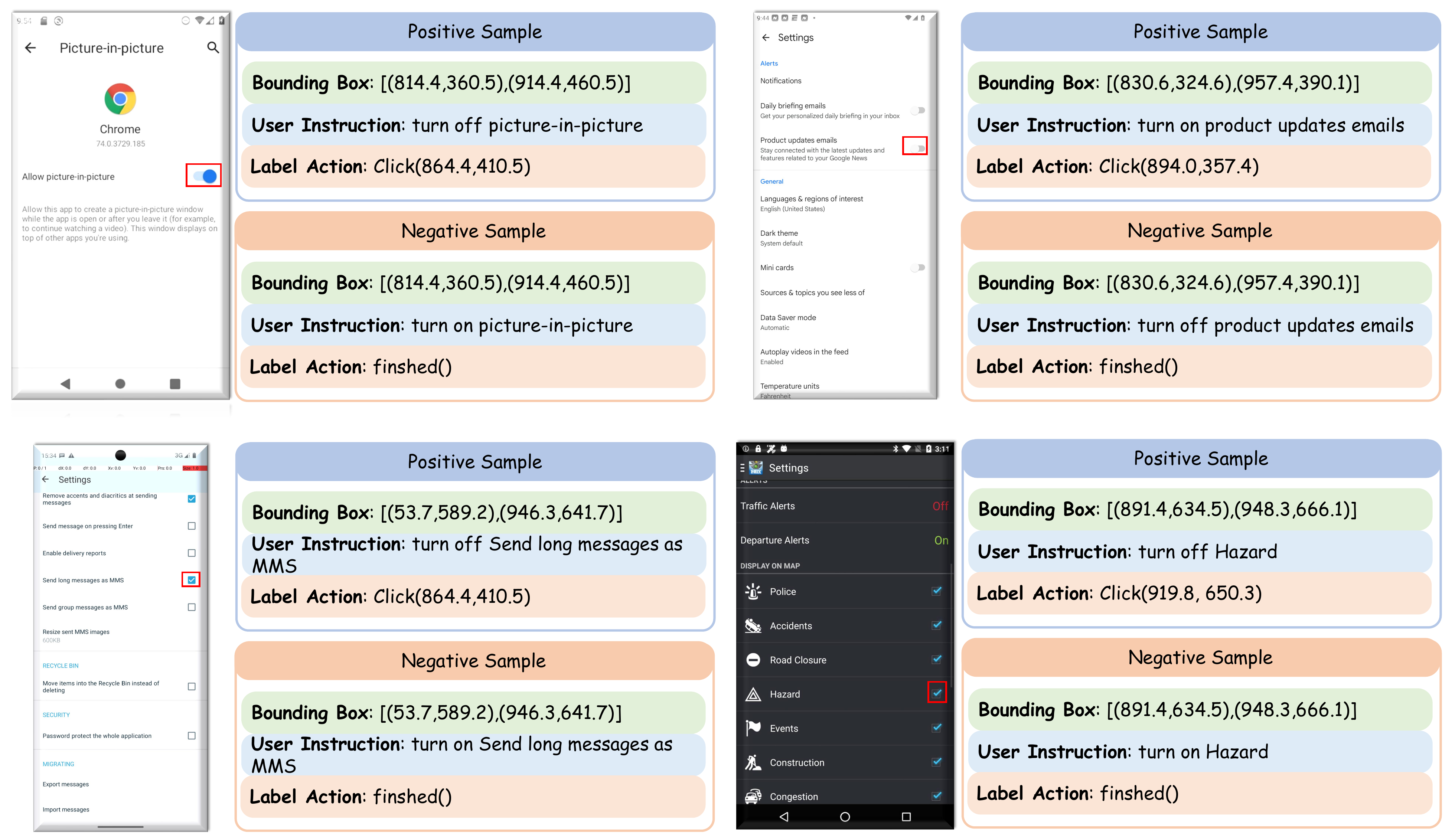}	\vspace{-0.4cm}
  \caption{Examples from the test split of state control benchmark, with target toggles highlighted in red boxes for clarity. \vspace{-0.2cm}}
  \label{fig:exampleFromStateControlBenchmarkTest}
\end{figure*}

\noindent (\romannumeral 1) \textbf{Widget Parsing.} Given that screenshots $s \in \mathbb{S}$ corresponding to toggle control instructions $u_t \in \mathbb{U}$ may contain more than one GUI toggle (as shown in Figure~\ref{fig:exampleFromStateControlBenchmarkTest}), we apply OminiParser~\cite{lu2024omniparser} to parse additional bounding boxes $b_p \in \mathbb{B}$ for clickable elements from these screenshots to enrich toggle diversity. We then merge the original and parsed bounding boxes into a unified set $\{b\} = \{b_o\} \cup \{b_p\}$, which serves as the basis for subsequent toggle identification.

\noindent (\romannumeral 2) \textbf{Toggle Identification.} Since the bounding boxes from the previous step may not always correspond to GUI toggles, we filter out non-toggle bounding boxes. Specifically, we adopt proprietary MLLMs, Qwen-2-VL-72B~\cite{wang2024qwen2} (denoted as $\mathcal{Q}$, \textbf{the best zero-shot baseline in Section~3.2}) and GLM-4V-Flash~\cite{glm2024chatglm} (denoted as $\mathcal{G}$, \textbf{freely available}), as independent annotators to recognize GUI toggles. Since proprietary MLLMs perform poorly in GUI grounding (as shown in Section~3.2), directly identifying toggles from bounding boxes is both challenging and inaccurate. Therefore, for each bounding box $b$ and its associated screenshot $s$, we highlight $b$ on $s$ to obtain box-highlighted screenshot $s_b$, which visually guides proprietary MLLMs to focus on the widget bounded by $b$ for more accurate widget recognition. Each annotator ($\mathcal{G}$ and $\mathcal{Q}$) serves as an indicator $\mathcal{I}$ to determine whether $b$ is a toggle based on box-highlighted screenshot $s_b$. The prompt template is provided in Appendix~\ref{appendix:prompts}. To ensure reliability, we apply inter-annotator agreement: only when both $\mathcal{G}$ and $\mathcal{Q}$ classify $b$ as a toggle do we retain $\langle s_b, b \rangle$. This process is formally defined as follows, where $\mathbf{1}[\cdot]$ is the indicator function that outputs 1 when the inner condition is met, otherwise outputs 0. 

\begin{equation}
  \label{equ:toggleIdentification_appendix}
  \begin{aligned}
  & \mathcal{I}_{\mathcal{G}}(s_b, b), \ \mathcal{I}_{\mathcal{Q}}(s_b, b) \in \{0,1\}, \\
  & \mathcal{I}_m(s_b, b) = m(s_b, b), \ m \in \{\mathcal{G}, \mathcal{Q}\}, \\
  & \mathcal{I}(s_b, b) = \mathbf{1}\left[\mathcal{I}_{\mathcal{G}}(s_b, b) = \mathcal{I}_{\mathcal{Q}}(s_b, b)\right].
  \end{aligned}
\end{equation}

\begin{table*}[!t]
  \centering
  \small
  \caption{Data source distribution for the state control benchmark.}
\label{tab:dataSourceDistribution}
\begin{tabular}{@{}ccccccccc@{}}
\toprule
Split & AITW~\cite{rawles2023android}  & RICOSCA~\cite{li2020widget} & OS-Atlas~\cite{wu2025osatlas} & AMEX~\cite{chai2024amex} & AndroidWorld~\cite{rawles2024androidworld} & GUIAct-Mobile~\cite{chen2025guicourse}  & Total \\ \midrule

Train & 68,380 & 4,144    & 496      & 444  & 130          & 58             & 73,652 \\
Test  & 7,558  & 496     & 60       & 56   & 12           & 2            & 8,184  \\ \bottomrule
\end{tabular}\vspace{-0.4cm}
\end{table*} 

\noindent (\romannumeral 3) \textbf{State-functionality Annotation.} This step employs $\mathcal{G}$ and $\mathcal{Q}$ as independent annotators to label the GUI toggle state and its functionality. Given the bounding box $b$ of a GUI toggle and the corresponding box-highlighted screenshot $s_b$, each annotator independently determines the current toggle state $\sigma$, where 0 indicates the toggle is currently off and 1 indicates it is on, and toggle functionality $f$. The prompt template for this annotation is provided in Appendix~\ref{appendix:prompts}. To ensure label reliability, we apply inter-annotator agreement: only when both $\mathcal{G}$ and $\mathcal{Q}$ produce identical annotations for both $\sigma$ and $f$, we accept the final annotation $\langle s_b, b, \sigma, f \rangle$. The process of state-functionality annotation is formally represented as follows.

\begin{equation}
  \label{equ:stateFeatureAnnotation_appendix}
  \begin{aligned}
  & \sigma_m(s_b, b), \ f_m(s_b, b) = m(s_b, b), \  m \in \{\mathcal{G}, \mathcal{Q}\}, \\
  & \mathcal{I}_{\sigma}(s_b, b) = \mathbf{1}\left[\sigma_{\mathcal{G}}(s_b, b) = \sigma_{\mathcal{Q}}(s_b, b)\right], \\
  & \mathcal{I}_{f}(s_b, b) = \mathbf{1}\left[f_{\mathcal{G}}(s_b, b) = f_{\mathcal{Q}}(s_b, b)\right].
  \end{aligned}
\end{equation}

After annotation, we obtain 40,918 screenshots.  For more comprehensive and practical evaluation, we replace the box-highlighted screenshots $s_b$ with the original screenshots $s$ and then expand each record with both positive and negative instructions based on the annotated toggle. Positive instructions $u_p$ require clicking the toggle to change the state, while negative ones $u_n$ require maintaining the current state. This yields 81,836 samples, which are then split into 73,652 balanced training and 8,184 testing samples, where each positive sample corresponds to a negative sample within the same split. The data source distribution for the state control benchmark is shown in Table~\ref{tab:dataSourceDistribution}.

Examples of test samples are provided in Figure~\ref{fig:exampleFromStateControlBenchmarkTest}, with target toggles highlighted in red boxes for clarity.

\noindent $\blacktriangleright$ \textbf{Data Quality.} Although proprietary MLLMs are not fully reliable for precise annotation, we apply two strategies to ensure the quality of the state control benchmark. First, we highlight the bounding box of each GUI toggle on the corresponding screenshot. This helps proprietary MLLMs focus on the widget for more accurate recognition, mitigating the limitation of low grounding ability. Second, we employ inter-annotator agreement to filter out inconsistent annotations. This strategy improves the reliability of the annotation. The details are as follows.

Assume two proprietary MLLMs $\mathcal{Q}$ and $\mathcal{G}$ act as independent annotators with probabilities $p_1$ and $p_2$ of correct annotation. Since we retain only matching annotations of the two annotators, the proportion $p$ of correct annotations among all retained annotations is given by:

\begin{equation}
  \label{equ:correctAnnotationProportion}
    p = \frac{p_1 p_2}{p_1 p_2 + (1 - p_1)(1 - p_2)}.
\end{equation}

\begin{table}[!t]
\centering
\small
\caption{Estimated probabilities of correct annotation $p_1$ and $p_2$ for two proprietary MLLMs, obtained by sampling 100 instances of state-functionality annotation and manually verifying their quality.}
\label{tab:annotationQuality}
\begin{tabular}{@{}ccc@{}}
\toprule
$p_1$ & $p_2$     & $p$ (Theoretical estimation) \\ \midrule
$0.81$ & $0.80$ & $0.946$  \\ \bottomrule
\end{tabular} 
\end{table}

As we highlight the bounding box of each GUI toggle to reduce grounding errors and improve recognition, $p_1$ and $p_2$ are greatly improved. We assume $p_1 > p_2$, then:

\begin{equation}
\label{equ:proofOfCorrectAnnotationProportion}
p - p_1 = \frac{p_1 (1 - p_1)(2 p_2 - 1)}
    {p_1 p_2 + (1 - p_1)(1 - p_2)} .
\end{equation}

The condition $p>p_1$ holds when $p_2 > 0.5$, indicating that in this case, $p$ exceeds both $p_1$ and $p_2$. We estimate $p_1$ and $p_2$ by sampling 100 instances of state-functionality annotation and manually checking the annotation quality, with results presented in Table~\ref{tab:annotationQuality}. Since $p_2 > 0.5$, the retained annotations through inter-annotator agreement are more reliable than those from a single proprietary MLLM. 


To assess the final benchmark quality, we manually verify 200 randomly sampled disjoint instances, finding that \textbf{92.5\% of functionality and 91\% of state annotations} match the ground truth, which aligns with our estimation in Table~\ref{tab:annotationQuality}. This confirms that our annotation pipeline ensures high annotation accuracy and overall benchmark reliability.

\noindent $\blacktriangleright$ \textbf{Metrics.} The comprehensive metric definitions for the state control benchmark are provided as follows. Let $\hat{t}_i$ and $t_i$ denote the predicted and ground truth action type of $i$-th sample, respectively. Let $\hat{a}_i$, $a_i$, $a_i^f$ represent the predicted action, ground truth action, and corresponding inverse toggle action, respectively. $\mathbf{1}[\cdot]$ denotes the indicator function. $\mathcal{P}$ and $\mathcal{N}$ denote the sets of positive and negative samples, respectively, and $N$ denotes the total number of samples.

\noindent (\romannumeral 1) \textbf{Overall Type Match Rate (O-TMR)} $\uparrow$: Proportion of test samples where the predicted action type (\texttt{CLICK} or \texttt{COMPLETED}) matches the ground truth. The formal definition of O-TMR is provided in Equation~\ref{equ:OTMR}.

\begin{equation}
  \label{equ:OTMR}
  \text{O-TMR} = \frac{1}{N} \sum_{i=1}^{N} \mathbf{1}\left[\hat{t}_i = t_i \right].
\end{equation}

\noindent (\romannumeral 2) \textbf{Overall Action Match Rate (O-AMR)} $\uparrow$: Proportion of test samples where the predicted action exactly matches the ground truth, considering both action type and click coordinate accuracy. O-AMR is the most critical metric on state control benchmark, reflecting the overall action precision of multimodal agents. The detailed evaluation process of action match rate is provided in Appendix~\ref{subappendix:evaluationOfAMR}. The formal definition of O-AMR is provided in Equation~\ref{equ:OAMR}.

\begin{equation}
  \label{equ:OAMR}
  \text{O-AMR} = \frac{1}{N} \sum_{i=1}^{N} \mathbf{1}\left[\hat{a}_i = a_i\right].
\end{equation}

\noindent (\romannumeral 3) \textbf{Positive Type Match Rate (P-TMR)} $\uparrow$: Proportion of positive samples where the predicted action type (\texttt{CLICK}) matches the ground truth. The formal definition of P-TMR is provided in Equation~\ref{equ:PTMR}.

\begin{equation}
  \label{equ:PTMR}
  \text{P-TMR} = \frac{1}{|\mathcal{P}|} \sum_{i \in \mathcal{P}} \mathbf{1}\left[\hat{t}_i = t_i\right].
\end{equation}

\noindent (\romannumeral 4) \textbf{Positive Action Match Rate (P-AMR)} $\uparrow$: Proportion of positive samples where the predicted action exactly matches the ground truth, considering both type and click coordinate accuracy. The formal definition of P-AMR is provided in Equation~\ref{equ:PAMR}.

\begin{equation}
  \label{equ:PAMR}
  \text{P-AMR} = \frac{1}{|\mathcal{P}|} \sum_{i \in \mathcal{P}} \mathbf{1}\left[\hat{a}_i = a_i\right].
\end{equation}

\noindent (\romannumeral 5) \textbf{Positive False Negative Rate (P-FNR)} $\downarrow$: Proportion of positive samples incorrectly predicted as negative (\texttt{COMPLETED}), reflecting the severity of false negatives. The formal definition of P-FNR is provided in Equation~\ref{equ:PFNR}.

\begin{equation}
  \label{equ:PFNR}
  \text{P-FNR} = \frac{1}{|\mathcal{P}|} \sum_{i \in \mathcal{P}} \mathbf{1}\left[\hat{t}_i = \texttt{COMPLETED}\right].
\end{equation}

\noindent (\romannumeral 6) \textbf{Negative Action Match Rate (N-AMR)} $\uparrow$: Proportion of negative samples where the predicted action (\texttt{COMPLETED}) matches the ground truth. The formal definition of N-AMR is provided in Equation~\ref{equ:NAMR}.

\begin{equation}
  \label{equ:NAMR}
  \text{N-AMR} = \frac{1}{|\mathcal{N}|} \sum_{i \in \mathcal{N}} \mathbf{1}\left[\hat{a}_i = a_i \right].
\end{equation}

\noindent (\romannumeral 7) \textbf{Negative False Positive Type Rate (N-FPTR)} $\downarrow$: Proportion of negative samples incorrectly predicted as \texttt{CLICK}, reflecting the false-positive tendency. The formal definition of N-FPTR is provided in Equation~\ref{equ:NFPTR}.

\begin{equation}
  \label{equ:NFPTR}
  \text{N-FPTR} = \frac{1}{|\mathcal{N}|} \sum_{i \in \mathcal{N}} \mathbf{1}\left[\hat{t}_i = \texttt{CLICK}\right].
\end{equation}

\noindent (\romannumeral 8) \textbf{Negative False Positive Rate (N-FPR)} $\downarrow$: Proportion of negative samples where the predicted \texttt{CLICK} coincides with the corresponding positive action, indicating the severity of false positives. The formal definition of N-FPR is provided in Equation~\ref{equ:NFPR}.

\begin{equation}
  \label{equ:NFPR}
  \text{N-FPR} = \frac{1}{|\mathcal{N}|} \sum_{i \in \mathcal{N}} \mathbf{1}\left[\hat{a}_i = a_i^f \ \land\ \hat{t}_i = \texttt{CLICK}\right].
\end{equation}

\subsection{Evaluation of Action Match Rate}\label{subappendix:evaluationOfAMR}

\begin{table*}[!t]
  \centering
  \small
\centering\caption{The action space with corresponding action parameters and descriptions in our experiments }
  \label{tab:actionSpace}
\begin{tabular}{@{}ccc@{}}
\toprule
Action   Type & Action Usage                    & Description                                           \\ \midrule
\texttt{CLICK}         & \texttt{CLICK} $[x,y]$                 & Click on the coordinate point $[x, y]$.     \\
\texttt{SCROLL}        & \texttt{SCROLL} [UP/DOWN/LEFT/RIGHT] & Scroll in the specified direction.                    \\
\texttt{TYPE}          & \texttt{TYPE} [content]              & Type the given content.                               \\
\texttt{OPENAPP}       & \texttt{OPENAPP} [app\_name]         & Open an app named [app\_name].                          \\
\texttt{COMPLETE}      & \texttt{COMPLETE}                        & Mark the current task as completed.                   \\
\texttt{WAIT}          & \texttt{WAIT}                            & Wait for the page or content to finish loading.       \\
\texttt{PRESS\_BACK}   & \texttt{PRESS\_BACK}                     & Press the back button to return to the previous page. \\
\texttt{PRESS\_HOME}   & \texttt{PRESS\_HOME}                     & Press the home button to return to the home screen.   \\
\texttt{PRESS\_ENTER}  & \texttt{PRESS\_ENTER}                    & Press the Enter key.                                  \\ \bottomrule
\end{tabular} \vspace{0.5cm}
\end{table*}

The exact action match rate (AMR) is a key metric for evaluating step-wise action prediction, as it requires both the action type $t$ and parameters $p$ (e.g., coordinates, app name, text input) to match the ground truth. The action space, along with corresponding parameters and descriptions in our experiments is provided in Table~\ref{tab:actionSpace}. An action is considered an exact match only when both $t$ and $p$ exactly match the ground truth. The calculation of AMR varies depending on the action type, as outlined below:

For actions without parameters (e.g., \texttt{COMPLETE}), evaluation depends only on action type $t$. AMR is equivalent to type match rate (TMR) for these actions.

For \texttt{SCROLL}, we evaluate both action type $t$ and scroll direction (UP, DOWN, LEFT, or RIGHT) to ensure they perfectly align with the ground truth.

For \texttt{TYPE}, we adopt a comparatively less stringent evaluation. After verifying that the predicted action type $t$ is \texttt{TYPE}, both the ground truth and predicted text are converted to lowercase and trimmed of leading and trailing spaces. The action is considered a match if the normalized predicted text exactly matches the normalized ground truth.

\begin{table*}[!t]
  \small
  \centering\caption{Statistical information for the test subsets of all three agentic benchmarks. \vspace{-0.2cm}}
  \label{tab:datasetDetail}

\begin{tabular}{@{}cccccccccc@{}}
\toprule
Benchmark      & \texttt{CLICK} & \texttt{COMPLETE} & \texttt{SCROLL} & \texttt{TYPE} & \texttt{OPENAPP} & \texttt{PRESS} & Others & Total Step & Trajectory  \\ \midrule
AndroidControl & 5074  & 1543     & 1211   & 632  & 608     & 343   & 576    & 9987       & 1543 \\
AITZ           & 2736  & 504      & 601    & 500  & /       & 265   & 118    & 4724       & 506  \\
GUI-Odyssey    & 16658 & 1572     & 2622   & 2666 & /       & 2044  & 89     & 25651      & 1666 \\ \bottomrule
\end{tabular}
  
\end{table*}

For \texttt{OPENAPP}, we also adopt a comparatively less stringent evaluation. This is due to ambiguity in app names (e.g., voice recorder-unrecorder vs. voice recorder) and inconsistencies between ground truth actions and the low-level instructions of AndroidControl (e.g. \texttt{OPENAPP} \textit{Flipsnack} vs. \textit{open the flipsnack magazine app}). Specifically, we first verify that the predicted action type $t$ is \texttt{OPENAPP}, then normalize all words in the predicted and ground truth app names by converting them into lowercase and applying stemming to reduce variations in tense and person. If either normalized app name is a substring of the other, the action is considered an exact match.

\begin{table*}[!t]
\small
\centering
\caption{Task name and user instruction templates of the dynamic evaluation benchmark.}
\label{tab:detailsOfDynamicEvaluationBenchmark}
\begin{tabular}{@{}cc@{}}
\toprule
Task Name                              & User Instruction Template                                               \\ \midrule
SystemBluetoothTurnOff                 & Turn bluetooth off.                                                \\
SystemBluetoothTurnOffVerify           & Turn bluetooth off.                                                \\
SystemBluetoothTurnOn                  & Turn bluetooth on.                                                 \\
SystemBluetoothTurnOnVerify            & Turn bluetooth on.                                                 \\
SystemWifiTurnOff                      & Turn wifi off.                                                     \\
SystemWifiTurnOffVerify                & Turn wifi off.                                                     \\
SystemWifiTurnOn                       & Turn wifi on.                                                      \\
SystemWifiTurnOnVerify                 & Turn wifi on.                                                      \\
TurnOffWifiAndTurnOnBluetooth          & Turn off WiFi, then enable bluetooth                               \\
TurnOnWifiAndOpenApp                   & Turn on Wifi, then open the \{app\_name\} app                      \\
TurnOnAlarm9AM                         & Trun on alarm at 9:00 AM.                                          \\
TurnOffAlarm9AM                        & Trun off alarm at 9:00 AM.                                         \\
TurnOnCaptionYoutube                   & Turn on captions in Youtube's settings.                            \\
TurnOffCaptionYoutube                  & Turn off captions in Youtube's settings.                           \\
TurnOnDoNotDisturb                     & Turn on Do not Disturb                                             \\
TurnOffDoNotDisturb                    & Turn off Do not Disturb                                            \\
TurnOnSaveAndFillPaymentMethodsChrome  & Turn on save and fill payment methods in Chrome's settings.        \\
TurnOffSaveAndFillPaymentMethodsChrome & Turn off save and fill payment methods in Chrome's settings.       \\
TurnOnAlwaysSecureConnChrome           & Turn on Always use the secure connections in Chrome's   settings.  \\
TurnOffAlwaysSecureConnChrome          & Turn off Always use the secure connections in Chrome's   settings. \\ \bottomrule
\end{tabular} \vspace{-0.2cm}

\end{table*}

For \texttt{CLICK} actions, we slightly modify the evaluation method from OS-Atlas~\cite{wu2025osatlas}, leveraging the availability of widget bounding boxes. Specifically, when both the predicted and ground truth action types $t$ are \texttt{CLICK}, we first inspect the corresponding screenshot layout to identify the bounding box $b$ containing the ground truth coordinates $[x, y]$. If such a box exists, the action is considered correct if the predicted coordinates $[\hat{x}, \hat{y}]$ fall within $b$; otherwise, we measure the relative distance. If no bounding box is found, correctness is determined solely by the relative distance $d = \sqrt{(x - \hat{x})^2 + (y - \hat{y})^2}$ between the predicted and ground truth coordinates. For the state control benchmark, the fine-grained nature of GUI toggles makes the commonly adopted 14\% threshold overly permissive, as even such deviations may lead to failed toggling. We therefore consider a toggle action correct only if the relative distance is below 4\% of the screen (i.e., $d \le 40$ in our normalized coordinate system). For agentic benchmarks, we maintain the commonly adopted 14\% threshold (i.e., $d \le 140$), as agentic tasks generally tolerate higher deviation.

\subsection{Details of Agentic Benchmark}\label{subappendix:detailsOfAgenticBenchmark}
The agentic benchmarks adopted in this paper are:

\noindent $\bullet$ AndroidControl~\cite{li2024effects} is a large-scale mobile agent benchmark comprising 15,283 demonstrations with step-wise instructions. Data are collected from human raters performing various tasks on 833 apps across 40 categories on Android devices. The training subset of AndroidControl includes 89,144 step-wise samples.

\noindent $\bullet$ AITZ~\cite{zhang2024android} is a mobile agent benchmark derived from a subset of AITW~\cite{rawles2023android} and annotated by GPT-4o~\cite{hurst2024gpt} for chain-of-action-thought (CoAT) components. It includes 2,504 operation trajectories across 18,643 steps, categorized into five domains: General, Install, GoogleApps, Single, and Web Shopping. The training subset of AITZ contains 13,919 step-wise samples.

\noindent $\bullet$ GUI-Odyssey~\cite{rawles2024androidworld} is a large-scale mobile benchmark for training and evaluating cross-app navigation agents on complex, long-chain tasks. It consists of 8,334 episodes from 6 mobile devices, covering 6 cross-app task types, 212 apps, and over 1,400 app combinations. The training subset of GUI-Odyssey includes 101,486 step-wise samples.

The complementary characteristics of these benchmarks enable comprehensive evaluation of agent capabilities across multiple dimensions: AndroidControl provides broad coverage and generalization across applications, AITZ provides explicit reasoning traces with detailed annotations, and GUI Odyssey emphasizes complex long horizon tasks. We present detailed statistics of the test subsets for all three benchmarks in Table~\ref{tab:datasetDetail}.

\subsection{Details of Dynamic Evaluation Benchmark}\label{subappendix:detailsOfDynamicEvaluationBenchmark}


To evaluate the real-world applicability of StaR, we construct a dynamic evaluation benchmark consisting of 20 real-world toggle control tasks selected from daily mobile usage scenarios. This benchmark assesses three key aspects: (\romannumeral 1) overall task execution accuracy, (\romannumeral 2) false positive toggling when the current toggle state already matches the desired state (Verify cases), and (\romannumeral 3) false negative toggling in normal operation scenarios. 

The benchmark is implemented on the Android emulator provided by AndroidStudio and built upon the AndroidWorld~\cite{rawles2025androidworld} framework. We design two types of evaluation scenarios for each toggle operation: normal execution cases and verification cases (marked with \textit{Verify} suffix) where the current toggle state already matches the desired state. Table~\ref{tab:detailsOfDynamicEvaluationBenchmark} presents all task names and user instruction templates, covering system settings (e.g., WiFi/Bluetooth), applications (Youtube/Chrome/Alarm), and composite tasks.


Following AndroidWorld~\cite{rawles2025androidworld}, we adopt overall task success rate ranging from $[0,1]$ as the primary metric, calculated as the ratio of successfully completed subtasks. Overall task success rate primary reflects overall task execution accuracy and reveal the severity of false positives and false negatives in this benchmark. For composite tasks like ``Turn off WiFi, then enable Bluetooth'', the task success rate reflects partial completion (e.g., 0.5 if one subtask fails). 

\subsection{Implementation Details of StaR} \label{subappendix:implementationtailsOfStaR}

Rather than prompting, we train the multimodal agents to explicitly learn the StaR reasoning process. To improve the abilities on toggle state control tasks, we include the training split of the state control benchmark and adjust the reasoning process of each sample into StaR style. Additionally, to preserve performance on general agentic tasks and to learn to apply StaR reasoning adaptively in toggle state control tasks, we also annotate commonly adopted agentic benchmarks. Specifically, for those episodes representing toggle state control tasks, we identify the concrete step of toggling and refine the corresponding reasoning process into StaR-style. For other steps, we insert the phase ``\textit{Target toggle not found in this screen}'' into the original reasoning process to help the agents learn to apply StaR reasoning only on critical toggling steps. For episodes that do not represent toggle state control tasks, we directly adopt the original reasoning process. After training on both the state control benchmark and the agentic benchmarks, the multimodal agents learn to apply StaR reasoning adaptively in the critical steps for toggling and to retain the original reasoning in other steps, improving toggle accuracy without sacrificing general agentic performance.

\begin{table}[!t]
\centering
\caption{Training hyperparameter settings of StaR.}
\label{tab:trainingHyperparametersOfStaR}
 \setlength{\tabcolsep}{15pt}
\begin{tabular}{@{}cc@{}}
\toprule
Hyperparameter                  & Value                      \\ \midrule
finetuning\_type                & full                       \\
freeze\_vision\_tower           & False                      \\
freeze\_multi\_modal\_projector & False                      \\
cutoff\_len                     & 8192                       \\
per\_device\_train\_batch\_size & 1                          \\
gradient\_accumulation\_steps   & 8                          \\
learning\_rate                  & $5 \times 10^{-6}$         \\
epochs                          & 3                          \\
lr\_scheduler\_type             & cosine                     \\
warmup\_ratio                   & 0.1                        \\
parameter\_data\_type           & bf16                       \\ \bottomrule
\end{tabular}
\end{table}

To train multimodal agents, we adopt their respective original prompt templates and click coordinate settings. For OS-Atlas-7B~\cite{wu2025osatlas}, UI-TARS-7B~\cite{qin2025ui}, and AgentCPM-GUI-8B~\cite{zhang2025agentcpmgui}, the click coordinates are normalized to $[0, 1000]$. For GUI-OWL-7B~\cite{ye2025mobile}, the click coordinates are set to original pixel coordinates. We adopt the LLaMA-Factory~\cite{zheng2024llamafactory} framework to train the multimodal agents, with detailed training hyperparameters provided in Table~\ref{tab:trainingHyperparametersOfStaR}. Additionally, FlashAttention~\cite{dao2023flashattention2} is adopted for acceleration.

For evaluation, as each multimodal agent has its own action format, we translate the action format into OS-Atlas-style~\cite{wu2025osatlas} for unified evaluation. For UI-TARS, which adopts multi screenshot history modeling, we evaluate each episode independently, adopting the previously predicted four steps within the same episode as the historical input. For general proprietary MLLM-based agents that lack dedicated GUI-agentic prompts, we adopt the UI-TARS prompt due to its simplicity. Similarly, we adopt the OS-Atlas prompt~\cite{wu2025osatlas} for Qwen-2-VL-72B~\cite{wang2024qwen2}. For GUI-R1~\cite{luo2025gui}, which is built on Qwen-2.5-VL-7B~\cite{bai2025qwen2}, we follow its original prompt configuration and set the click coordinates to original pixel coordinates.

\begin{table*}[!t]
  \centering
  \small
  \setlength{\tabcolsep}{6pt}
\caption{Detailed evaluation results of existing multimodal agents on state control benchmark. Subscripts denote absolute improvements over the zero-shot baseline, with red indicating improvements and green indicating degradations. The optimal and the suboptimal results are \textbf{bolded} and \underline{underlined}, respectively. Results demonstrate that StaR training significantly improves execution and grounding accuracy on state control benchmark, outperforming prompting baselines significantly and highlighting the necessity of training. \vspace{-0.2cm}}
\label{tab:detailedEvaluationResultsOnStateControlBenchmark}
\begin{tabular}{@{}ccccccccc@{}}
\toprule
Model          & O-TMR$\uparrow$ & O-AMR$\uparrow$ & P-TMR$\uparrow$ & P-AMR$\uparrow$ & P-FNR$\downarrow$ & N-AMR$\uparrow$ & N-FPTR$\downarrow$ & N-FPR$\downarrow$ \\ \midrule 
\rowcolor{gray!20}
\multicolumn{9}{c}{\textit{Zero-shot}}                                                   \\ \midrule
GPT-5          & 75.35 & 37.05 & 91.91 & 15.30 & 2.79  & 58.80 & 36.14  & \underline{3.01}  \\
GPT-4o         & 72.40 & 27.17 & 97.04 & 6.60  & 2.35  & 47.75 & 48.83  & \textbf{2.39}  \\
Gemini-2.5-Pro & 68.74 & 30.25 & \textbf{98.85} & 21.87 & \textbf{0.78}  & 38.64 & 60.14  & 9.31  \\
Qwen-2-VL-72B  & \textbf{87.59} & \textbf{66.42} & 96.21 & \underline{53.89} & 3.69  & \textbf{78.96} & \textbf{20.67}  & {6.33}  \\
GUI-R1-7B         & 78.27 & 54.14 & {97.58} & 49.32 & 2.03  & 58.97 & 40.37  & 12.63 \\
OS-Atlas-7B       & 67.16 & 43.95 & \underline{98.51} & 52.10 & \underline{1.27}  & 35.80 & 64.10  & 28.67 \\
UI-TARS-7B & 67.14 & 47.45 & 94.33 & 54.94 & {1.71} & 39.96 & 48.29 & 17.62 
\\
AgentCPM-GUI-8B   & \underline{81.74} & \underline{64.08} & 95.38 & \textbf{60.04} & 3.32  & \underline{68.11} & \underline{30.69}  & {11.07} \\ 
GUI-Owl-7B & 76.58 & 53.57 & 94.99 & 48.97 & 2.32 & 58.16 & 39.15 & 14.66 
 \\ 
\midrule
\rowcolor{gray!20}
\multicolumn{9}{c}{\textit{w/ State-focused Prompt Engineering}} \\ \midrule

GPT-5          & $82.09_{\textcolor{darkred}{\uparrow 6.74}}$ & $46.35_{\textcolor{darkred}{\uparrow 9.30}}$ & $87.32_{\textcolor{darkgreen}{\downarrow 4.59}}$ & $15.84_{\textcolor{darkred}{\uparrow 0.54}}$ & $4.25_{\textcolor{darkgreen}{\uparrow 1.46}}$ & $76.86_{\textcolor{darkred}{\uparrow 18.06}}$ & $\textbf{15.00}_{\textcolor{darkred}{\downarrow 21.14}}$ & $\underline{1.22}_{\textcolor{darkred}{\downarrow 1.79}}$   \\

GPT-4o         & $82.87_{\textcolor{darkred}{\uparrow 10.47}}$ & $38.66_{\textcolor{darkred}{\uparrow 11.49}}$ & $94.01_{\textcolor{darkgreen}{\downarrow 3.03}}$ & $5.60_{\textcolor{darkgreen}{\downarrow 1.00}}$ & $4.79_{\textcolor{darkgreen}{\uparrow 2.44}}$ & $71.73_{\textcolor{darkred}{\uparrow 23.98}}$ & $25.88_{\textcolor{darkred}{\downarrow 22.95}}$ & $\textbf{1.03}_{\textcolor{darkred}{\downarrow 1.36}}$   \\

Gemini-2.5-Pro & $81.78_{\textcolor{darkred}{\uparrow 13.04}}$ & $42.86_{\textcolor{darkred}{\uparrow 12.61}}$ & $\textbf{97.29}_{\textcolor{darkgreen}{\downarrow 1.56}}$ & $19.45_{\textcolor{darkgreen}{\downarrow 2.42}}$ & $\textbf{2.39}_{\textcolor{darkgreen}{\uparrow 1.61}}$ & $66.28_{\textcolor{darkred}{\uparrow 27.64}}$ & $33.06_{\textcolor{darkred}{\downarrow 27.08}}$ & $3.74_{\textcolor{darkred}{\downarrow 5.57}}$   \\

Qwen-2-VL-72B & $\underline{87.11}_{\textcolor{darkgreen}{\downarrow 0.48}}$ & $\underline{65.29}_{\textcolor{darkgreen}{\downarrow 0.36}}$ & ${95.85}_{\textcolor{darkgreen}{\downarrow 1.69}}$ & $52.20_{\textcolor{darkred}{\uparrow 0.34}}$ & $4.03_{\textcolor{darkgreen}{\uparrow 0.34}}$ & $\underline{78.37}_{\textcolor{darkgreen}{\downarrow 0.59}}$ & $20.99_{\textcolor{darkgreen}{\uparrow 0.32}}$ & $7.87_{\textcolor{darkgreen}{\uparrow 1.54}}$ \\

GUI-R1-7B & $\textbf{89.15}_{\textcolor{darkred}{\uparrow 10.88}}$ & $\textbf{65.59}_{\textcolor{darkred}{\uparrow 11.45}}$ & $94.47_{\textcolor{darkgreen}{\downarrow 3.11}}$ & $47.65_{\textcolor{darkgreen}{\downarrow 1.67}}$ & $4.99_{\textcolor{darkgreen}{\uparrow 2.96}}$ & $\textbf{83.53}_{\textcolor{darkred}{\uparrow 24.56}}$ & $\underline{16.13}_{\textcolor{darkred}{\downarrow 24.24}}$ & $4.37_{\textcolor{darkred}{\downarrow 8.26}}$ \\

OS-Atlas-7B  &    $75.21_{\textcolor{darkred}{\uparrow 8.05}}$ & $52.55_{\textcolor{darkred}{\uparrow 8.60}}$ & $94.53_{\textcolor{darkgreen}{\downarrow 3.98}}$ & $49.22_{\textcolor{darkgreen}{\downarrow 2.88}}$ & $4.42_{\textcolor{darkgreen}{\uparrow 3.15}}$ & $55.89_{\textcolor{darkred}{\uparrow 20.09}}$ & $43.57_{\textcolor{darkred}{\downarrow 20.53}}$ & $18.91_{\textcolor{darkred}{\downarrow 9.76}}$   \\

UI-TARS-7B & $81.84_{\textcolor{darkred}{\uparrow 14.70}}$ & $63.23_{\textcolor{darkred}{\uparrow 15.78}}$ & $93.38_{\textcolor{darkgreen}{\downarrow 0.95}}$ & $\underline{56.16}_{\textcolor{darkred}{\uparrow 1.22}}$ & $5.91_{\textcolor{darkgreen}{\uparrow 4.20}}$ & $70.31_{\textcolor{darkred}{\uparrow 30.35}}$ & $27.83_{\textcolor{darkred}{\downarrow 20.46}}$ & $9.38_{\textcolor{darkred}{\downarrow 8.24}}$ \\

AgentCPM-GUI-8B & $82.30_{\textcolor{darkred}{\uparrow 0.63}}$ & $64.67_{\textcolor{darkred}{\uparrow 0.59}}$ & $95.43_{\textcolor{darkred}{\uparrow 0.05}}$ & $\textbf{60.04}_{\textcolor{darkgreen}{\downarrow 0.00}}$& ${3.47}_{\textcolor{darkgreen}{\uparrow 0.15}}$ & $69.31_{\textcolor{darkred}{\uparrow 1.20}}$ & $30.01_{\textcolor{darkred}{\downarrow 0.68}} $& $10.65_{\textcolor{darkred}{\downarrow 0.42}}$   \\ 

GUI-Owl-7B & $85.02_{\textcolor{darkred}{\uparrow 8.44}}$ & $61.09_{\textcolor{darkred}{\uparrow 7.52}}$ & $\underline{96.02}_{\textcolor{darkred}{\uparrow 1.03}}$ & $48.17_{\textcolor{darkgreen}{\downarrow 0.80}}$ & $\underline{3.05}_{\textcolor{darkgreen}{\uparrow 0.73}}$ & $74.02_{\textcolor{darkred}{\uparrow 15.86}}$ & $24.51_{\textcolor{darkred}{\downarrow 14.64}}$ & $6.96_{\textcolor{darkred}{\downarrow 7.70}}$ \\

\midrule

\rowcolor{gray!20}
\multicolumn{9}{c}{\textit{w/ StaR-style Prompting}} \\ \midrule
Qwen-2-VL-72B & $\textbf{87.81}_{\textcolor{darkred}{\uparrow 0.22}}$ & $\textbf{65.91}_{\textcolor{darkgreen}{\downarrow 0.51}}$ & $95.50_{\textcolor{darkgreen}{\downarrow 0.71}}$ & $51.71_{\textcolor{darkgreen}{\downarrow 2.18}}$ & $4.37_{\textcolor{darkgreen}{\uparrow 0.68}}$ & $\underline{80.11}_{\textcolor{darkred}{\uparrow 1.15}}$ & $\underline{19.67}_{\textcolor{darkred}{\downarrow 1.00}}$ & $\underline{6.33}_{\textcolor{darkgreen}{\uparrow 0.00}}$ \\

GUI-R1 & $\underline{87.77}_{\textcolor{darkred}{\uparrow 9.50}}$ & $\underline{65.10}_{\textcolor{darkred}{\uparrow 10.96}}$ & $\underline{95.58}_{\textcolor{darkgreen}{\downarrow 2.00}}$ & $48.24_{\textcolor{darkgreen}{\downarrow 1.08}}$ & $4.18_{\textcolor{darkgreen}{\uparrow 2.15}}$ & $\textbf{81.96}_{\textcolor{darkred}{\uparrow 22.99}}$ & $\textbf{17.77}_{\textcolor{darkred}{\downarrow 22.60}}$ & $\textbf{5.47}_{\textcolor{darkred}{\downarrow 7.16}}$ \\

OS-Atlas-7B       & $73.52_{\textcolor{darkred}{\uparrow 6.36}}$   &        $50.07_{\textcolor{darkred}{\uparrow 6.12}}$   &        $\textbf{96.77}_{\textcolor{darkgreen}{\downarrow 1.74}}$   &        $49.88_{\textcolor{darkgreen}{\downarrow 2.22}}$   &        $\textbf{2.96}_{\textcolor{darkgreen}{\uparrow 1.69}}$   &        $50.27_{\textcolor{darkred}{\uparrow 14.47}}$   &        $49.62_{\textcolor{darkred}{\downarrow 14.48}}$   &        $22.21_{\textcolor{darkred}{\downarrow 6.46}}$
  \\

UI-TARS-7B & $81.18_{\textcolor{darkred}{\uparrow 14.04}}$ & $62.89_{\textcolor{darkred}{\uparrow 15.44}}$ & $90.98_{\textcolor{darkgreen}{\uparrow 3.35}}$ & $\underline{54.40}_{\textcolor{darkgreen}{\uparrow 0.54}}$ & $8.55_{\textcolor{darkgreen}{\uparrow 6.84}}$ & ${71.38}_{\textcolor{darkred}{\uparrow 31.42}}$ & ${27.54}_{\textcolor{darkred}{\downarrow 20.75}}$ & ${9.38}_{\textcolor{darkred}{\downarrow 8.24}}$ \\

AgentCPM-GUI-8B   & ${82.14}_{\textcolor{darkred}{\uparrow 0.40}}$   &        ${64.43}_{\textcolor{darkred}{\uparrow 0.35}}$   &        ${95.36}_{\textcolor{darkgreen}{\downarrow 0.02}}$   &        $\textbf{59.95}_{\textcolor{darkgreen}{\downarrow 0.09}}$   &        $\underline{3.59}_{\textcolor{darkgreen}{\uparrow 0.27}}$   &        $68.91_{\textcolor{darkred}{\uparrow 0.80}}$   &        $30.28_{\textcolor{darkred}{\downarrow 0.41}}$   &        $10.58_{\textcolor{darkred}{\downarrow 0.49}} $ \\ 

GUI-Owl-7B & $84.27_{\textcolor{darkred}{\uparrow 7.69}}$ & $60.92_{\textcolor{darkred}{\uparrow 7.35}}$ & $94.06_{\textcolor{darkgreen}{\downarrow 0.93}}$ & $47.36_{\textcolor{darkgreen}{\downarrow 1.61}}$ & $4.55_{\textcolor{darkgreen}{\uparrow 2.23}}$ & ${74.49}_{\textcolor{darkred}{\uparrow 16.33}}$ & ${23.68}_{\textcolor{darkred}{\downarrow 15.47}}$ & ${7.48}_{\textcolor{darkred}{\downarrow 7.18}}$
 \\
\midrule

\rowcolor{gray!20}
\multicolumn{9}{c}{\textit{w/ Ground Truth Toggle State Prompting}} \\ \midrule
Qwen-2-VL-72B & $\textbf{94.00}_{\textcolor{darkred}{\uparrow 6.41}}$ & $\textbf{72.14}_{\textcolor{darkred}{\uparrow 5.72}}$ & $95.80_{\textcolor{darkgreen}{\downarrow 0.41}}$ & $52.08_{\textcolor{darkgreen}{\downarrow 1.81}}$ & $4.13_{\textcolor{darkgreen}{\uparrow 0.44}}$ & $\textbf{92.20}_{\textcolor{darkred}{\uparrow 13.24}}$ & $\textbf{7.70}_{\textcolor{darkred}{\downarrow 12.97}}$ & $\textbf{2.37}_{\textcolor{darkred}{\downarrow 3.96}}$ 
 \\

GUI-R1 & $\underline{85.26}_{\textcolor{darkred}{\uparrow 6.99}}$ & $60.63_{\textcolor{darkred}{\uparrow 6.49}}$ & $\underline{97.63}_{\textcolor{darkred}{\uparrow 0.05}}$ & $48.36_{\textcolor{darkgreen}{\downarrow 0.96}}$ & $1.98_{\textcolor{darkred}{\downarrow 0.05}}$ & $\underline{72.90}_{\textcolor{darkred}{\uparrow 13.93}}$ & $\underline{26.78}_{\textcolor{darkred}{\downarrow 13.59}}$ & $\underline{7.87}_{\textcolor{darkred}{\downarrow 4.76}}$ \\

OS-Atlas-7B &  $68.29_{\textcolor{darkred}{\uparrow 1.13}}$ & $45.22_{\textcolor{darkred}{\uparrow 1.27}}$ & $\textbf{98.46}_{\textcolor{darkgreen}{\downarrow 0.05}}$ & $52.32_{\textcolor{darkred}{\uparrow 0.22}}$ & $\textbf{1.34}_{\textcolor{darkgreen}{\uparrow 0.07}}$ & $38.12_{\textcolor{darkred}{\uparrow 2.32}}$ & $61.73_{\textcolor{darkred}{\downarrow 2.37}}$ & $28.15_{\textcolor{darkred}{\downarrow 0.52}}$ \\

UI-TARS-7B & $73.53_{\textcolor{darkred}{\uparrow 6.39}}$ & $54.03_{\textcolor{darkred}{\uparrow 6.58}}$ & $96.11_{\textcolor{darkred}{\uparrow 1.78}}$ & $\underline{57.18}_{\textcolor{darkred}{\uparrow 2.24}}$ & $\underline{1.81}_{\textcolor{darkgreen}{\uparrow 0.10}}$ & $50.88_{\textcolor{darkred}{\uparrow 10.92}}$ & $42.67_{\textcolor{darkred}{\downarrow 5.62}}$ & $15.18_{\textcolor{darkred}{\downarrow 2.44}}$ \\

AgentCPM-GUI-8B   & $82.48_{\textcolor{darkred}{\uparrow 0.74}}$ & $\underline{64.88}_{\textcolor{darkred}{\uparrow 0.80}}$ & $95.58_{\textcolor{darkred}{\uparrow 0.20}}$ & $\textbf{60.39}_{\textcolor{darkred}{\uparrow 0.35}}$ & $3.27_{\textcolor{darkred}{\downarrow 0.05}}$ & $69.38_{\textcolor{darkred}{\uparrow 1.27}}$ & $29.99_{\textcolor{darkred}{\downarrow 0.70}}$ & $10.63_{\textcolor{darkred}{\downarrow 0.44}}$ 
 \\ 

GUI-Owl-7B & $83.22_{\textcolor{darkred}{\uparrow 6.64}}$ & $60.07_{\textcolor{darkred}{\uparrow 6.50}}$ & $96.68_{\textcolor{darkred}{\uparrow 1.69}}$ & $50.37_{\textcolor{darkred}{\uparrow 1.40}}$ & $1.91_{\textcolor{darkred}{\downarrow 0.41}}$ & $69.77_{\textcolor{darkred}{\uparrow 11.61}}$ & $28.18_{\textcolor{darkred}{\downarrow 10.97}}$ & $9.19_{\textcolor{darkred}{\downarrow 5.47}}$ 
\\

\midrule

\rowcolor{gray!20}
\multicolumn{9}{c}{\textit{w/ Ground Truth Toggle State and StaR-style Prompting}} \\ \midrule
Qwen-2-VL-72B & $\textbf{92.85}_{\textcolor{darkred}{\uparrow 5.26}}$ & $\textbf{70.88}_{\textcolor{darkred}{\uparrow 4.46}}$ & $96.21_{\textcolor{darkgreen}{\downarrow 0.00}}$ & $52.27_{\textcolor{darkgreen}{\downarrow 1.62}}$ & $3.59_{\textcolor{darkred}{\downarrow 0.10}}$ & $\textbf{89.49}_{\textcolor{darkred}{\uparrow 10.53}}$ & $\textbf{10.26}_{\textcolor{darkred}{\downarrow 10.41}}$ & $\textbf{2.86}_{\textcolor{darkred}{\downarrow 3.47}}$
 \\

GUI-R1 &  $\underline{90.24}_{\textcolor{darkred}{\uparrow 11.97}}$ & $\underline{66.31}_{\textcolor{darkred}{\uparrow 12.17}}$ & $95.38_{\textcolor{darkgreen}{\downarrow 2.20}}$ & $47.53_{\textcolor{darkgreen}{\downarrow 1.79}}$ & $4.45_{\textcolor{darkgreen}{\uparrow 2.42}}$ & $\underline{85.09}_{\textcolor{darkred}{\uparrow 26.12}}$ & $\underline{14.81}_{\textcolor{darkred}{\downarrow 25.56}}$ & $\underline{4.57}_{\textcolor{darkred}{\downarrow 8.06}}$ 
\\

OS-Atlas-7B & $75.55_{\textcolor{darkred}{\uparrow 8.39}}$ & $51.78_{\textcolor{darkred}{\uparrow 7.83}}$ & $\underline{97.78}_{\textcolor{darkgreen}{\downarrow 0.73}}$ & $50.24_{\textcolor{darkgreen}{\downarrow 1.86}}$ & $\underline{2.15}_{\textcolor{darkgreen}{\uparrow 0.88}}$ & $53.32_{\textcolor{darkred}{\uparrow 17.52}}$ & $46.60_{\textcolor{darkred}{\downarrow 17.50}}$ & $19.97_{\textcolor{darkred}{\downarrow 8.70}}$  \\

UI-TARS-7B & $84.04_{\textcolor{darkred}{\uparrow 16.90}}$ & $65.29_{\textcolor{darkred}{\uparrow 17.84}}$ & $95.41_{\textcolor{darkred}{\uparrow 1.08}}$ & $\underline{57.89}_{\textcolor{darkred}{\uparrow 2.95}}$ & $4.08_{\textcolor{darkgreen}{\uparrow 2.37}}$ & $72.68_{\textcolor{darkred}{\uparrow 32.72}}$ & $26.56_{\textcolor{darkred}{\downarrow 21.73}}$ & $8.75_{\textcolor{darkred}{\downarrow 8.87}}$ 
\\

AgentCPM-GUI-8B   & $82.20_{\textcolor{darkred}{\uparrow 0.46}}$ & $64.37_{\textcolor{darkred}{\uparrow 0.29}}$ & $95.50_{\textcolor{darkred}{\uparrow 0.12}}$ & $\textbf{59.85}_{\textcolor{darkgreen}{\downarrow 0.19}}$ & $3.35_{\textcolor{darkgreen}{\uparrow 0.03}}$ & $68.89_{\textcolor{darkred}{\uparrow 0.78}}$ & $30.40_{\textcolor{darkred}{\downarrow 0.29}}$ & $10.78_{\textcolor{darkred}{\downarrow 0.29}}$ 
 \\ 

GUI-Owl-7B & $78.29_{\textcolor{darkred}{\uparrow 1.71}}$ & $54.70_{\textcolor{darkred}{\uparrow 1.13}}$ & $\textbf{98.00}_{\textcolor{darkred}{\uparrow 3.01}}$ & $50.83_{\textcolor{darkred}{\uparrow 1.86}}$ & $\textbf{1.44}_{\textcolor{darkred}{\downarrow 0.88}}$ & $58.58_{\textcolor{darkred}{\uparrow 0.42}}$ & $40.57_{\textcolor{darkgreen}{\uparrow 1.42}}$ & $15.66_{\textcolor{darkgreen}{\uparrow 1.00}}$ 
 \\

\midrule

\rowcolor{gray!20}
\multicolumn{9}{c}{\textit{w/ StaR Training}}                                                     \\ \midrule
OS-Atlas-7B       & $\textbf{96.13}_{\textcolor{darkred}{\uparrow 28.97}}$ & $\textbf{79.72}_{\textcolor{darkred}{\uparrow 35.77}}$ & $\textbf{95.77}_{\textcolor{darkgreen}{\downarrow 2.74}}$ & $\textbf{62.95}_{\textcolor{darkred}{\uparrow 10.85}}$ & $\textbf{4.23}_{\textcolor{darkgreen}{\uparrow 2.96}}$  & $96.48_{\textcolor{darkred}{\uparrow 60.68}}$ & $3.52_{\textcolor{darkred}{\downarrow 60.58}}$   & $1.52_{\textcolor{darkred}{\downarrow 27.15}}$  \\

UI-TARS-7B & $95.82_{\textcolor{darkred}{\uparrow 28.68}}$ & $77.86_{\textcolor{darkred}{\uparrow 30.41}}$ & ${95.11}_{\textcolor{darkred}{\uparrow 0.78}}$ & $59.19_{\textcolor{darkred}{\uparrow 4.25}}$ & $\underline{4.89}_{\textcolor{darkgreen}{\uparrow 3.18}}$ & $\underline{96.53}_{\textcolor{darkred}{\uparrow 56.57}}$ & $\underline{3.45}_{\textcolor{darkred}{\downarrow 44.84}}$ & ${1.34}_{\textcolor{darkred}{\downarrow 16.28}}$ \\

AgentCPM-GUI-8B   & $\underline{95.98}_{\textcolor{darkred}{\uparrow 14.24}}$ & $\underline{79.00}_{\textcolor{darkred}{\uparrow 14.92}}$ & $94.50_{\textcolor{darkgreen}{\downarrow 0.88}}$ & $\underline{60.53}_{\textcolor{darkred}{\uparrow 0.49}}$ & $5.50_{\textcolor{darkgreen}{\uparrow 2.18}}$  & $\textbf{97.46}_{\textcolor{darkred}{\uparrow 29.35}}$ & $\textbf{2.54}_{\textcolor{darkred}{\downarrow 28.15}}$   & $\textbf{0.95}_{\textcolor{darkred}{\downarrow 10.12}}$  \\ 

GUI-Owl-7B & $\underline{95.99}_{\textcolor{darkred}{\uparrow 19.41}}$ & $77.60_{\textcolor{darkred}{\uparrow 24.03}}$ & $\underline{95.65}_{\textcolor{darkred}{\uparrow 0.66}}$ & $58.87_{\textcolor{darkred}{\uparrow 9.90}}$ & $\textbf{4.35}_{\textcolor{darkgreen}{\uparrow 2.03}}$ & $96.33_{\textcolor{darkred}{\uparrow 38.17}}$ & $3.67_{\textcolor{darkred}{\downarrow 35.48}}$ & $\underline{1.56}_{\textcolor{darkred}{\downarrow 13.10}}$ \\

\bottomrule
\end{tabular}

\end{table*}

\section{Additional Results and Analyses}\label{appendix:additionalExperimentalResults}
This section reports additional experimental results and analyses. Section~\ref{subappendix:detailedEvaluationResultsOfExistingMultimodalAgentsOnStateControlBenchmark} provides detailed evaluations of existing multimodal agents on the state control benchmark with more comprehensive comparisons to four training free baselines, highlighting the effectiveness of StaR and the necessity of training. Section~\ref{subappendix:additionalPerformanceOnAgenticBenchmarks} reports results and analysis for omitted agents on agentic benchmarks, demonstrating that StaR training improves or maintains performance on general agentic tasks. Section~\ref{subappendix:ablationStudiesOnStaRComponents} presents ablation studies on StaR components, indicating that StaR is most effective when all three components are integrated. Section~\ref{subappendix:caseStudies} provides case studies illustrating how StaR improves perception, reasoning, and execution in toggle control tasks.

\subsection{Detailed Evaluation Results of Existing Multimodal Agents on State Control Benchmark}\label{subappendix:detailedEvaluationResultsOfExistingMultimodalAgentsOnStateControlBenchmark}


We present the detailed evaluation results of existing multimodal agents on the state control benchmark under different settings in Table~\ref{tab:detailedEvaluationResultsOnStateControlBenchmark}. Specifically, in addition to the vanilla zero-shot baseline, we include four training-free baselines: 

\noindent (\romannumeral 1) \textbf{State-focused Prompt Engineering,} where we append additional prompt to the original prompt to guide the agents to focus on toggle state during reasoning (see Section~3.2), with the prompt template in Appendix~\ref{appendix:prompts}.

\noindent (\romannumeral 2) \textbf{StaR-style Prompting}, where we append additional prompt to the original prompt to guide the agents to follow the StaR reasoning process (see Section~5.2), with the prompt template in Appendix~\ref{appendix:prompts}.

\noindent (\romannumeral 3) \textbf{Ground Truth Toggle State Prompting}, where we provide the ground truth current toggle state in the prompt, representing the \textbf{theoretical upper bound} of performance that a multi-agent collaboration framework achieves when an additional annotator agent identifies the toggle state with perfect accuracy and guides the reasoning of the action agent. The prompt template is provided in Appendix~\ref{appendix:prompts}.

\noindent (\romannumeral 4) \textbf{Ground Truth Toggle State and StaR-style Prompting}, where we provide the ground truth current toggle state in the prompt and append additional prompt to the original prompt to guide the agents to follow the StaR reasoning process. This baseline reflects the \textbf{theoretical upper bound} of performance achievable by a multi-agent collaboration framework that introduces an additional annotator agent to identify the toggle state with perfect accuracy and guide the reasoning of the action agent, along with following the StaR reasoning process. In other words, this baseline can reflect the \textbf{theoretical upper bound without training}.

\begin{figure*}[!t]
  \centering
  \begin{subfigure}{\linewidth}
    \centering
    \includegraphics[width=\linewidth]{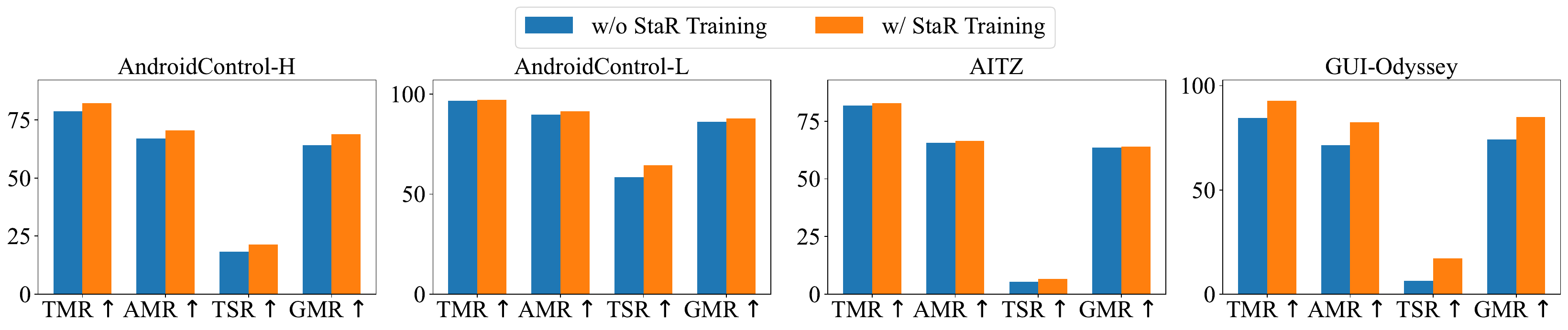}
    \caption{Performance of OS-Atlas-7B}
  \end{subfigure}
  
  \begin{subfigure}{\linewidth}
    \centering
    \includegraphics[width=\linewidth]{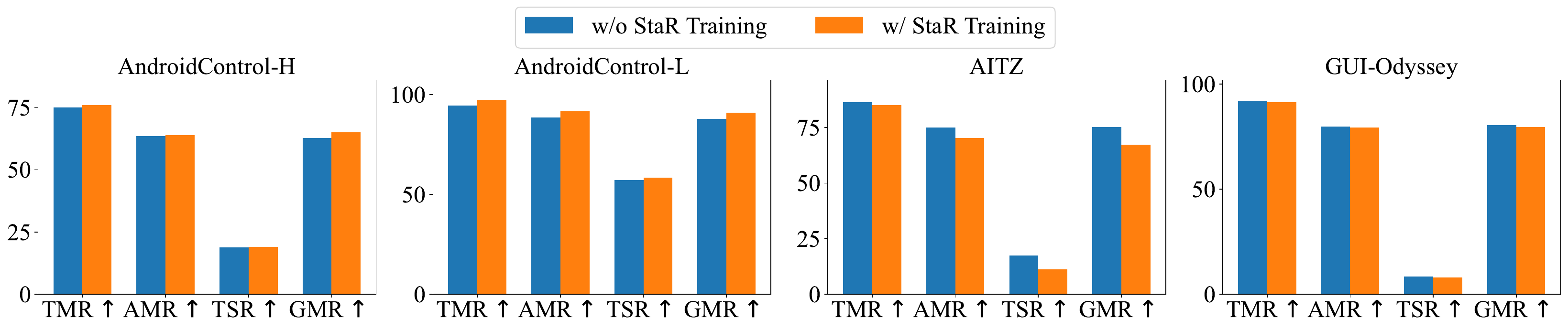}
    \caption{Performance of AgentCPM-GUI-8B}
  \end{subfigure}

  \begin{subfigure}{\linewidth}
    \centering
    \includegraphics[width=\linewidth]{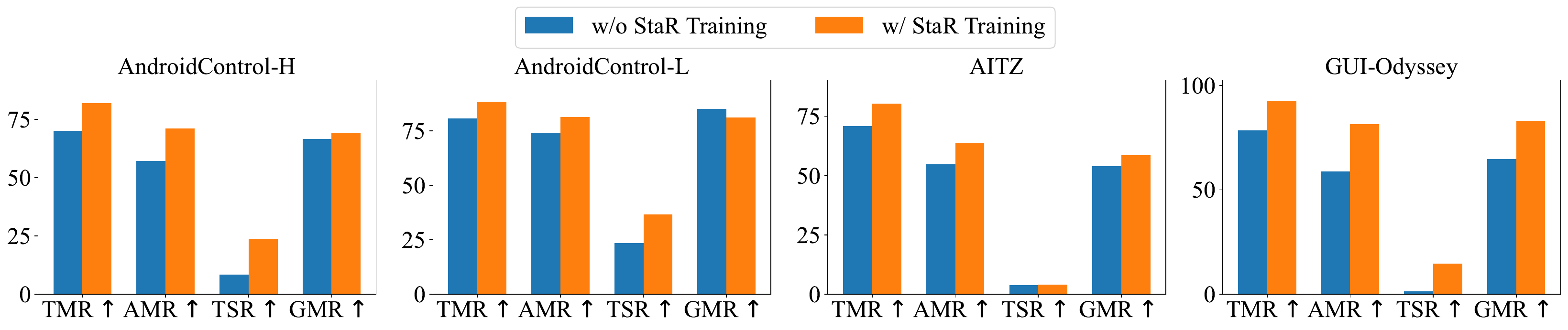}
    \caption{Performance of GUI-OWL-7B}
  \end{subfigure}

  \caption{The performance o f zero-shot and StaR-trained (a) OS-Atlas-7B; (b) AgentCPM-GUI-8B; and (c) GUI-OWL-7B on agentic benchmarks. Results demonstrate that StaR consistently preserves or enhances performance on agentic benchmarks.} \vspace{0.2cm}
  \label{fig:perfromanceOfOtherAgentsOnAgenticBenchmarks}
\end{figure*}


Based on the results, we draw the following conclusions:

(\romannumeral 1) Existing multimodal agents perform poorly in state control tasks. Even the best-performing agent yield less than 70\% O-AMR, and most agents yield less than 50\% O-AMR, which is close to random toggling. While these agents perform better on positive samples, their grounding ability remains limited, with P-AMR generally below 60\%. For negative samples, all agents show a strong bias toward false positive toggling, resulting in high N-FPTR. These results demonstrate the limited capability of current multimodal agents in state control.

(\romannumeral 2) All baselines without training do not fundamentally improve performance. For all baselines, weak agents such as OS-Atlas-7B and UI-TARS-7B achieves notable improvements on negative samples, but their overall performance remains unsatisfactory. For strong agents such as Qwen-2-VL-72B and AgentCPM-GUI-8B, the improvements are marginal and may even result in degradations. This is likely attributed to two reasons. On one hand, prompting does not fundamentally improve the grounding ability of multimodal agents, and the recognition of GUI toggles remains poor. On the other hand, prompting is less effective in establishing the mapping between negative instructions and the corresponding correct actions, resulting in improved but still unsatisfactory performance on negative samples. Collectively, These results suggest that relying solely on prompting or multi-agent annotation collaboration without training is not sufficient to address the limitations of multimodal agents in state control.

\begin{table*}[!t]
  \centering
  \small
  \setlength{\tabcolsep}{2pt}
\caption{Ablation results of OS-Atlas-7B with different StaR training components, where P, A, D represent perceiving, analyzing, and deciding, respectively. Subscripts denote absolute improvements over the vanilla baseline, with red indicating improvements and green indicating degradations. The optimal results are \textbf{bolded}. Removing perceiving or analyzing consistently degrades performance, confirming that StaR is most effective when all three components are integrated. }
\label{tab:ablationOfStaROnAtlas}
\begin{tabular}{@{}cccccccccccc@{}}
\toprule
\multirow{2}{*}{Method} & \multicolumn{3}{c}{Components}             & \multirow{2}{*}{O-TMR$\uparrow$} & \multirow{2}{*}{O-AMR$\uparrow$} & \multirow{2}{*}{P-TMR$\uparrow$} & \multirow{2}{*}{P-AMR$\uparrow$} & \multirow{2}{*}{P-FNR$\downarrow$} & \multirow{2}{*}{N-AMR$\uparrow$} & \multirow{2}{*}{N-FPTR$\downarrow$} & \multirow{2}{*}{N-FPR$\downarrow$} \\
                        & P   & A    & D     &                                  &                                  &                                  &                                  &                                    &                                  &                                     &                                    \\ \midrule
Vanilla                 & $\times$     & $\times$     & $\times$     & 67.16                            & 43.95                            & 98.51                            & 52.10                            & 1.27                               & 35.80                            & 64.10                               & 28.67                              \\ \midrule
\makecell{StaR w/o Perceiving}     & $\times$     & $\checkmark$ & $\checkmark$ & $89.97_{\textcolor{darkred}{\uparrow 22.81}}$                            & $73.39_{\textcolor{darkred}{\uparrow 29.44}}$                            & $95.94_{\textcolor{darkgreen}{\downarrow 2.57}}$                            & $62.78_{\textcolor{darkred}{\uparrow 10.68}}$                            & $\textbf{4.06}_{\textcolor{darkgreen}{\uparrow 2.79}}$                               & $83.99_{\textcolor{darkred}{\uparrow 48.19}}$                            & $16.01_{\textcolor{darkred}{\downarrow 48.09}}$                               & $9.41_{\textcolor{darkred}{\downarrow 19.26}}$                               \\
\makecell{StaR w/o Analyzing}      & $\checkmark$ & $\times$     & $\checkmark$ & $90.57_{\textcolor{darkred}{\uparrow 23.41}}$                            & $73.94_{\textcolor{darkred}{\uparrow 29.99}}$                            & $95.63_{\textcolor{darkgreen}{\downarrow 2.88}}$                            & $62.32_{\textcolor{darkred}{\uparrow 10.22}}$                            & $4.35_{\textcolor{darkgreen}{\uparrow 3.08}}$                               & $85.56_{\textcolor{darkred}{\uparrow 49.76}}$                            & $14.42_{\textcolor{darkred}{\downarrow 49.68}}$                               & $9.92_{\textcolor{darkred}{\downarrow 18.75}}$                               \\
StaR                    & $\checkmark$ & $\checkmark$ & $\checkmark$ & $\textbf{96.13}_{\textcolor{darkred}{\uparrow 28.97}}$                            & $\textbf{79.72}_{\textcolor{darkred}{\uparrow 35.77}}$                            & $\textbf{95.77}_{\textcolor{darkgreen}{\downarrow 2.74}}$                            & $\textbf{62.95}_{\textcolor{darkred}{\uparrow 10.85}}$                            & $4.23_{\textcolor{darkgreen}{\uparrow 2.96}}$                               & $\textbf{96.48}_{\textcolor{darkred}{\uparrow 60.68}}$                            & $\textbf{3.52}_{\textcolor{darkred}{\downarrow 60.58}}$                                & $\textbf{1.52}_{\textcolor{darkred}{\downarrow 27.15}}$                               \\ \bottomrule


\end{tabular}

\end{table*}

(\romannumeral 3) Only StaR training leads to substantial improvements, highlighting the necessity of training. StaR training consistently improves both O-AMR and P-AMR, surpassing that of all baselines significantly. For positive samples, StaR training can yield improvements on P-AMR, likely due to training on toggle clicks that strengthen toggle recognition. In contrast, P-AMR generally decreases under prompting-based baselines. For negative samples, StaR training achieves nearly 100\% N-AMR, significantly mitigating false positive toggling and surpassing all baselines. These findings highlight the effectiveness of StaR training in improving performance on both positive and negative samples, reinforcing the necessity of training to learn this state-aware structured reasoning process.

Collectively, StaR training demonstrates its effectiveness in improving the execution and grounding capabilities of multimodal agents on the state control benchmark.

\subsection{Detailed Performance on Agentic Benchmarks}\label{subappendix:additionalPerformanceOnAgenticBenchmarks}
We present the detailed performance of the omitted agents on the agentic benchmarks in Figure~\ref{fig:perfromanceOfOtherAgentsOnAgenticBenchmarks}.

From the results, we have the following findings:

(\romannumeral 1) StaR consistently improves or preserves general agentic performance. Across four benchmark settings, StaR training consistently preserves or surpasses baselines. While AgentCPM-GUI-8B shows outlier degradations on AITZ and GUI-Odyssey, our error analysis on 463 degradation and 287 improvement cases before and after StaR training reveals that 28\% of degradations are due to coordinate drift and 37\% to valid alternative operations (e.g., clicking the back icon vs. PRESS\_BACK). As alternatives appear in both cases, coordinate drift is the primary cause. The drift likely stems from the fixed visual tokens of AgentCPM, which are less robust than the flexible tokens of Qwen-2-VL to the large resolution span of AITZ and GUI-Odyssey. Thus, this outlier reflects model-specific grounding sensitivity distinct from StaR effectiveness.


(\romannumeral 2) StaR generalizes across multimodal agents. Results on all agents indicate that StaR consistently improves performance on agentic benchmarks. These findings demonstrate that StaR is model-agnostic and can effectively enhance the reasoning ability of diverse multimodal agents.

Overall, the results indicate that StaR training does not compromise model capability. Additionally, in several settings, StaR leads to measurable improvements, highlighting its generalizability on general agentic tasks.

\subsection{Ablation Studies on StaR components}\label{subappendix:ablationStudiesOnStaRComponents}

To further evaluate the impact of each component (perceiving, analyzing, deciding) in the StaR reasoning chain for toggle control, we select OS-Atlas-7B as the target agent and train it with different combinations of these components. Since decision is generally equivalent to low-level instructions, we only consider removing perceiving or analyzing. The ablation results are presented in Table~\ref{tab:ablationOfStaROnAtlas}. 

From the results, we observe the following:

(\romannumeral 1) Removing perceiving or analyzing consistently degrades performance. Additionally, Ground Truth Toggle State Prompting in Appendix~\ref{subappendix:detailedEvaluationResultsOfExistingMultimodalAgentsOnStateControlBenchmark} can be approximately regarded as making action decisions based solely on perception. The corresponding results are still inferior to those of StaR training, which integrates all three reasoning components. These findings confirm that StaR is most effective when all three components are integrated. 

(\romannumeral 2) StaR training, even with some components removed, still outperforms the vanilla baseline, demonstrating its effectiveness in enabling precise toggle control.

\subsection{Case Studies}\label{subappendix:caseStudies}

\begin{figure*}[!t]
  \centering
  \includegraphics[width=\linewidth]{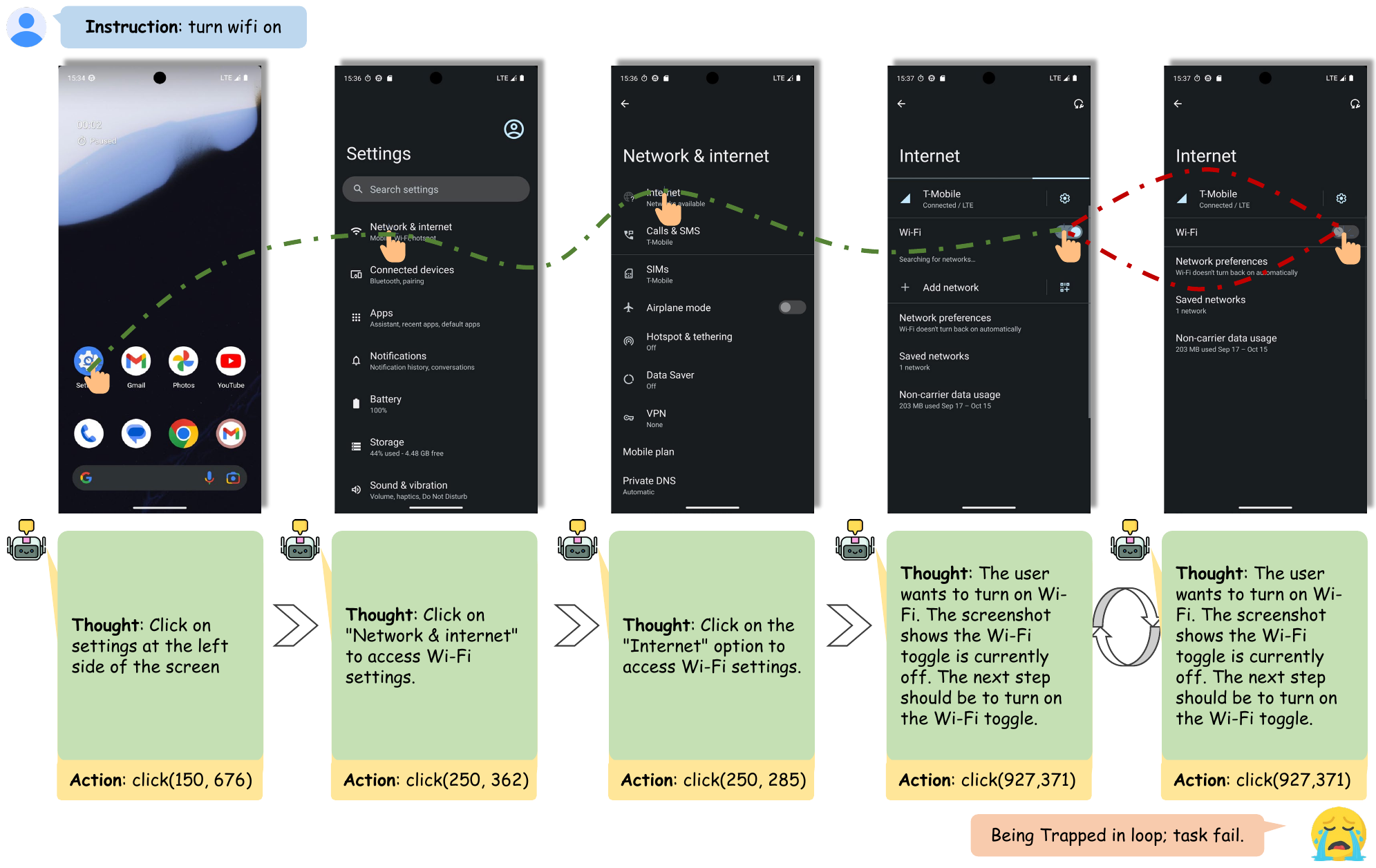}	
  \caption{Trajectory of OS-Atlas-7B without StaR training when executing the instruction \textit{turn wifi on} with the toggle initially \textit{on}. The agent mistakenly perceives the current toggle state as \textit{off} and incorrectly clicks the toggle, resulting in an unintended state change. It then repeatedly toggles between \textit{on} and \textit{off}, falling into an infinite loop and ultimately failing the task.}
  \label{fig:caseStudyOSAtlasWithoutStaR}
\end{figure*}

\begin{figure*}[!t]
  \centering
  \includegraphics[width=\linewidth]{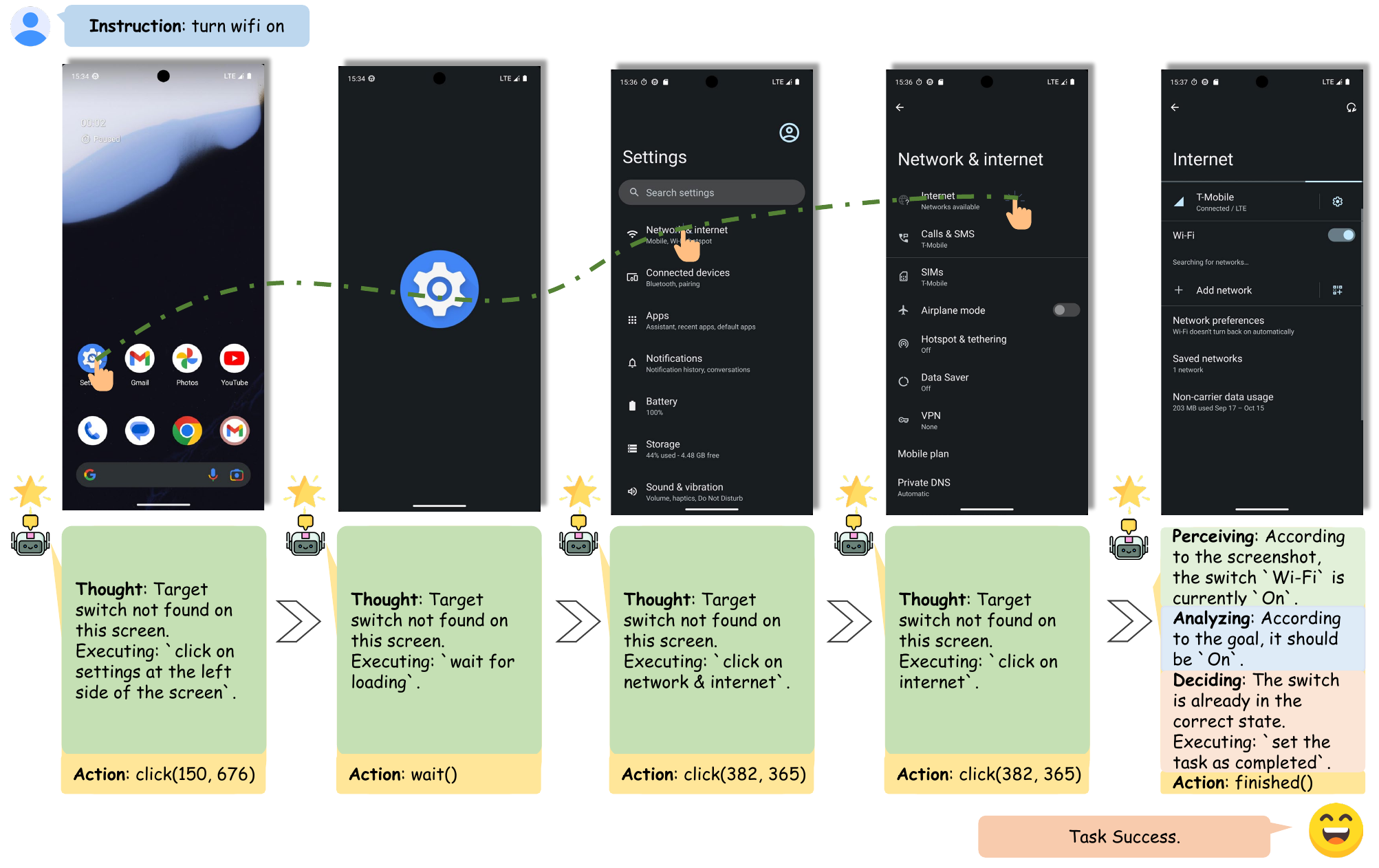}	
  \caption{Trajectory of OS-Atlas-7B with StaR training when executing the instruction \textit{turn wifi on} with the toggle initially \textit{on}. At the critical decision step, the agent adaptively applies the state-aware reasoning chain, correctly perceiving the current toggle state as \textit{on} and appropriately deciding to finish the task, thereby completing the instruction as intended.}
  \label{fig:caseStudyOSAtlasWithStaR}
\end{figure*}

To demonstrate the effectiveness of StaR-trained agents in precisely executing real-world toggle control instructions, we adopt OS-Atlas-7B, which exhibited the most pronounced improvement in Section~5.4, as a representative example. The instruction is ``\textit{turn wifi on}'', with the toggle initially set to \textit{on}, thereby testing false-positive toggling. The trajectories of OS-Atlas-7B without and with StaR training are presented in Figure~\ref{fig:caseStudyOSAtlasWithoutStaR} and Figure~\ref{fig:caseStudyOSAtlasWithStaR}, respectively.

From these examples, we observe the following:

(i) Without StaR training, OS-Atlas-7B fails to execute the instruction correctly, resulting in a false positive toggle. The agent misperceives the current toggle state as \textit{off} and incorrectly clicks the toggle, resulting in an unintended state change. It then repeatedly toggles between \textit{on} and \textit{off}, falling into an infinite loop and ultimately failing the task.

(ii) With StaR training, OS-Atlas-7B successfully executes the instruction correctly. At the critical decision step, the agent adaptively applies the state-aware reasoning chain, correctly perceiving the current toggle state as \textit{on} and appropriately deciding to finish the task, thereby completing the instruction as intended.

These case studies illustrate the effectiveness of StaR in enabling agents to precisely execute real-world toggle control instructions in dynamic environments.

\section{Prompts}\label{appendix:prompts}
\begin{figure*}[!t]
  \centering
  \includegraphics[width=\linewidth]{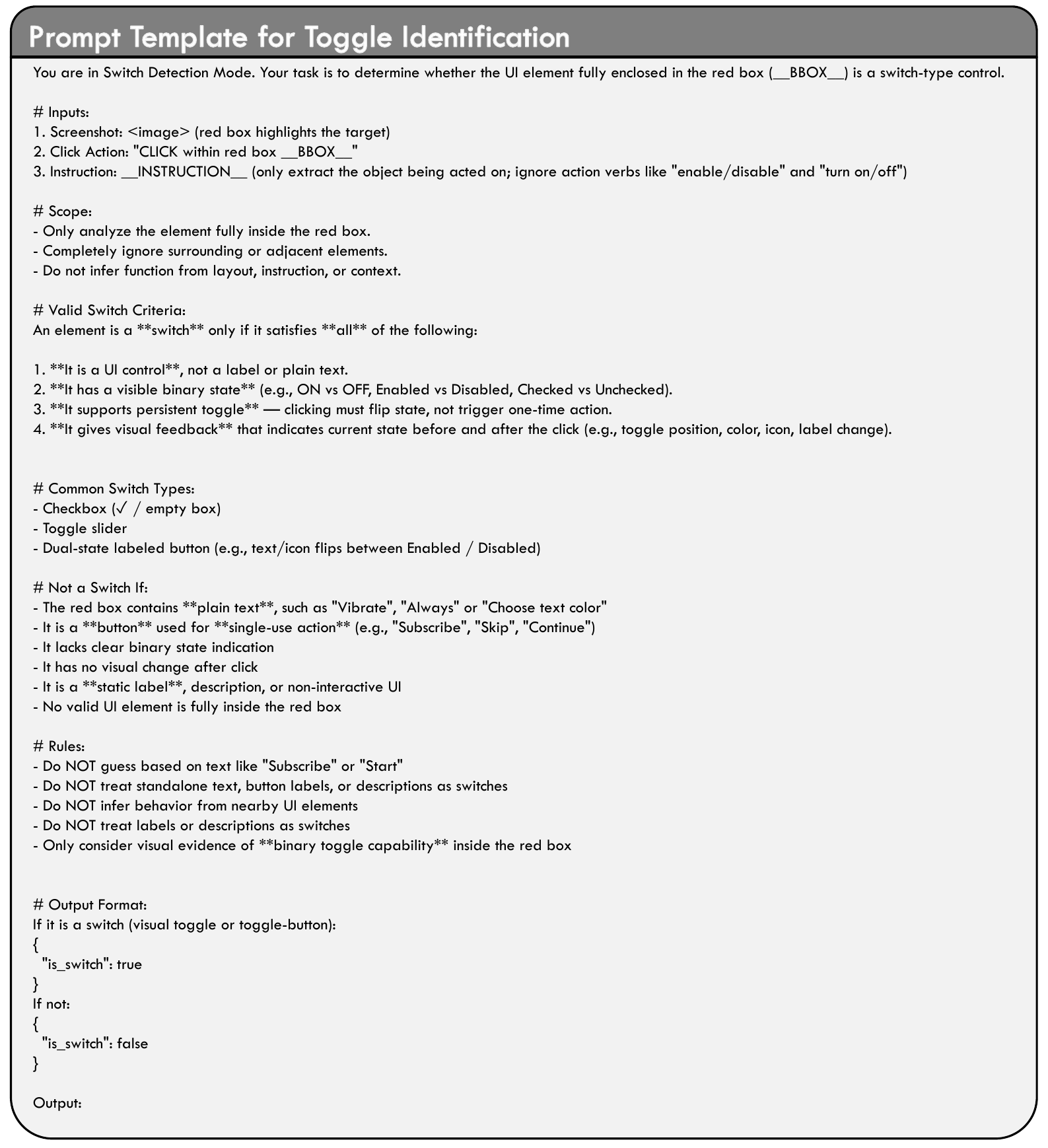}	
  \caption{Prompt template for toggle identification.}
  \label{fig:promptToggleIdentification}
\end{figure*}

\begin{figure*}[!t]
  \centering
  \includegraphics[width=\linewidth]{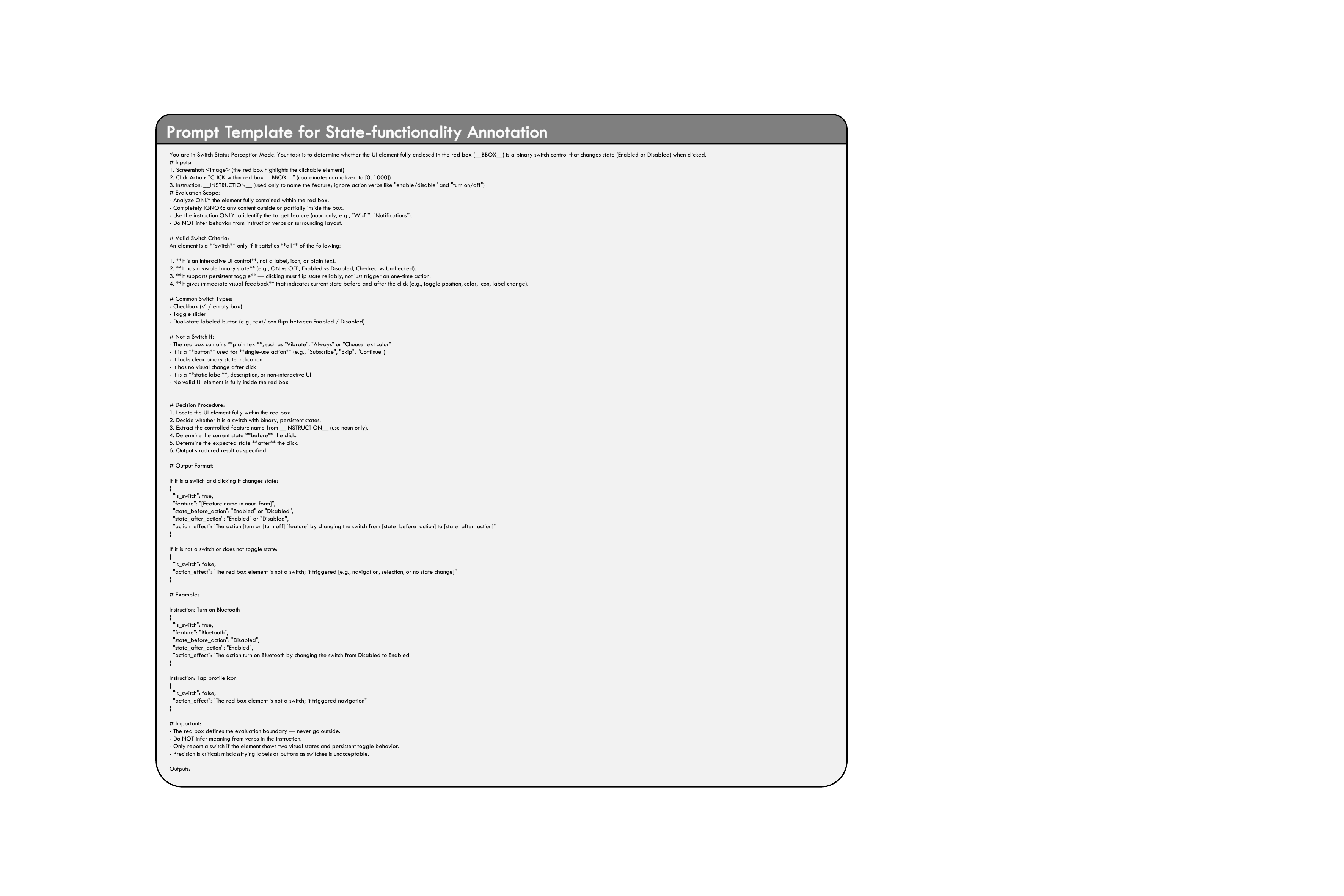}	
  \caption{Prompt template for state-functionality annotation.}
  \label{fig:promptStateFeatureAnnotation}
\end{figure*}

\begin{figure*}[!t]
  \centering
  \includegraphics[width=\linewidth]{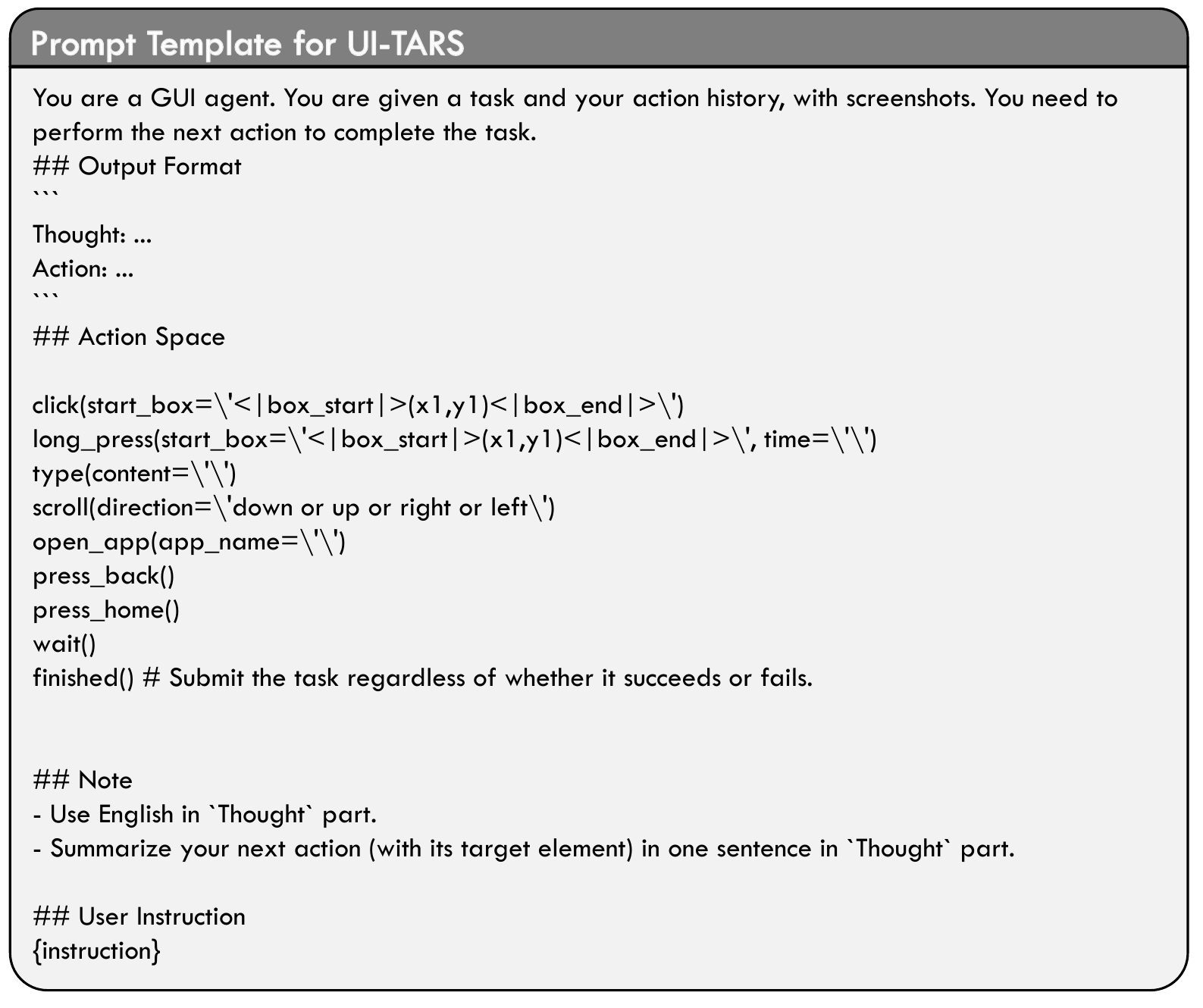}	
  \caption{Prompt template for UI-TARS.}
  \label{fig:promptUITARS}
\end{figure*}

\begin{figure*}[!t]
  \centering
  \includegraphics[width=\linewidth]{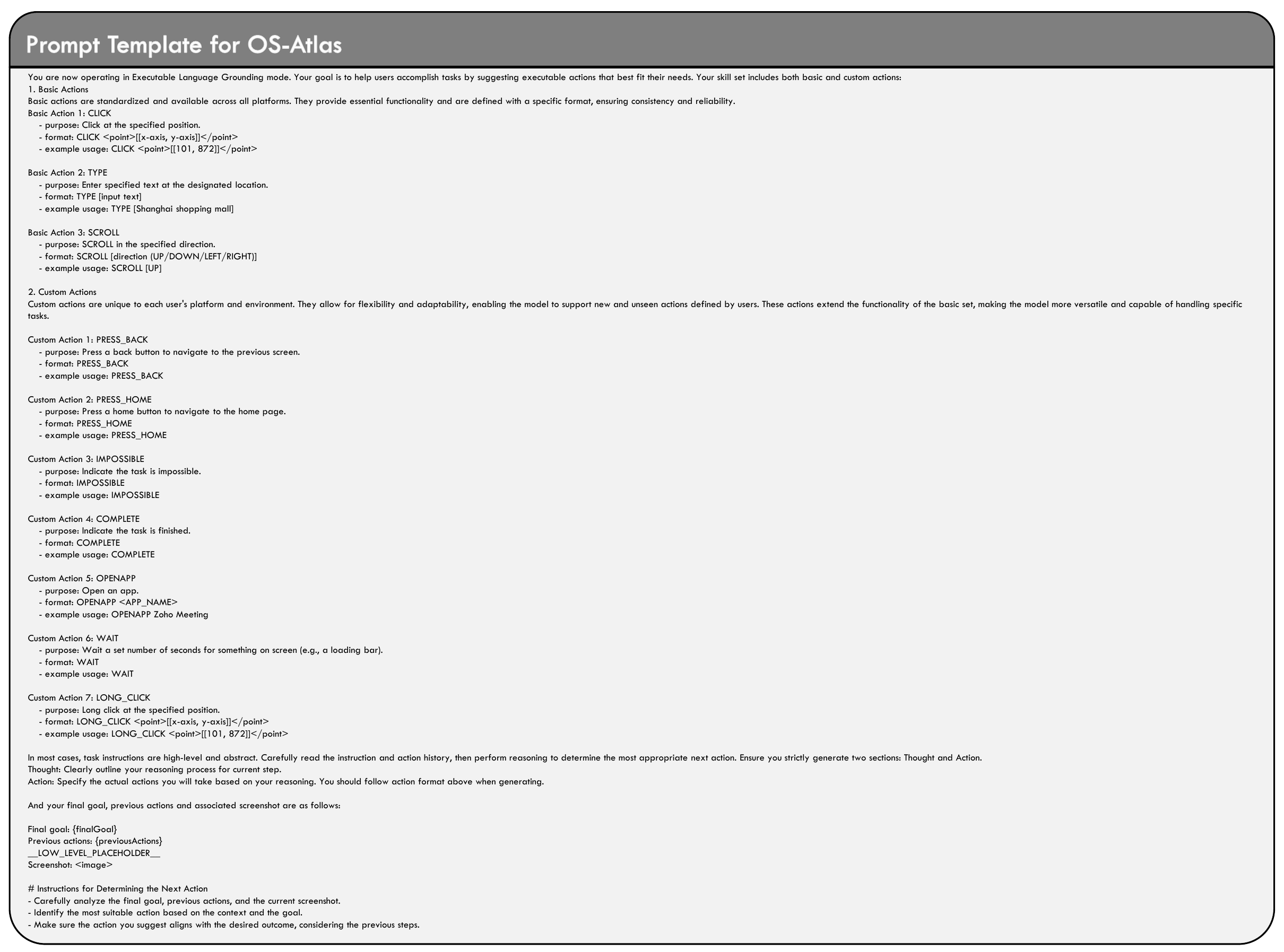}	
  \caption{Prompt template for OS-Atlas.}
  \label{fig:promptOSATlas}
\end{figure*}

\begin{figure*}[!t]
  \centering
  \includegraphics[width=\linewidth]{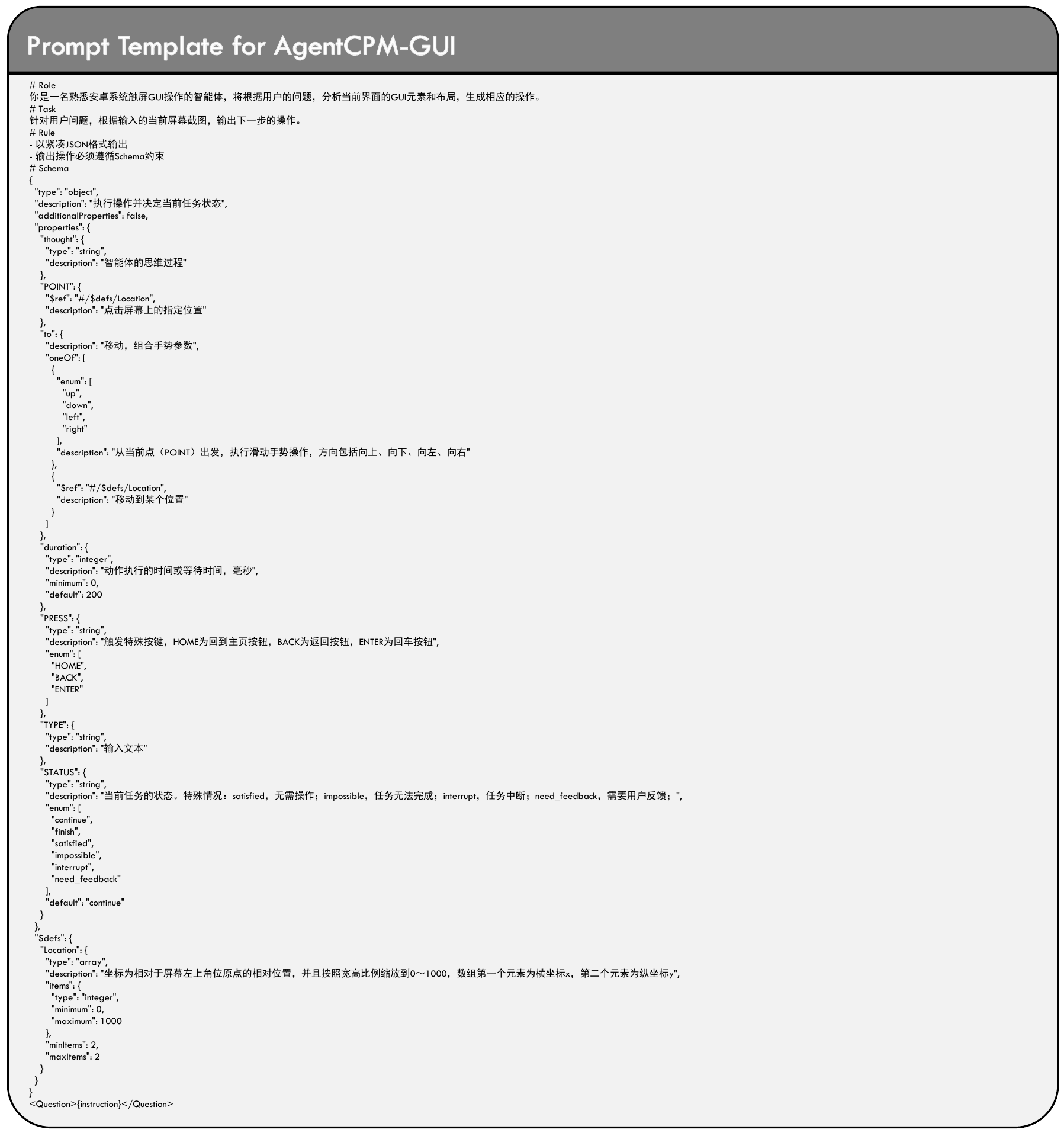}	
  \caption{Prompt template for AgentCPM-GUI.}
  \label{fig:promptAgentCPMGUI}
\end{figure*}

\begin{figure*}[!t]
  \centering
  \includegraphics[width=\linewidth]{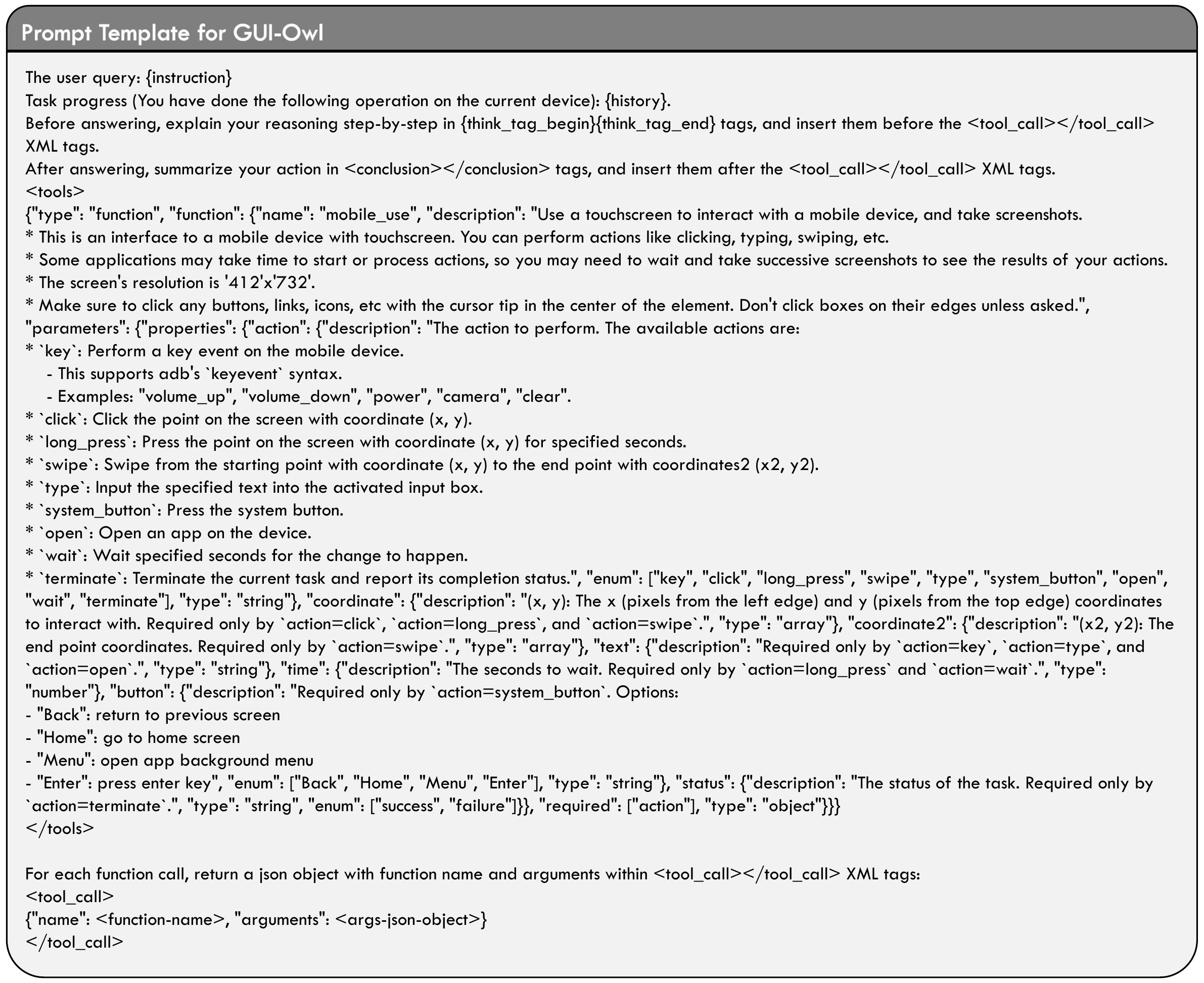}	
  \caption{Prompt template for GUI-Owl.}
  \label{fig:promptGUIOwl}
\end{figure*}

This section presents the meticulously designed prompt templates. The template for toggle identification, state-functionality annotation, UI-TARS, OS-Atlas, AgentCPM-GUI, and GUI-Owl are provided in Figure~\ref{fig:promptToggleIdentification}, Figure~\ref{fig:promptStateFeatureAnnotation}, Figure~\ref{fig:promptUITARS}, Figure~\ref{fig:promptOSATlas}, Figure~\ref{fig:promptAgentCPMGUI}, and Figure~\ref{fig:promptGUIOwl}, respectively. 

\begin{figure*}[!t]
  \centering
  \includegraphics[width=\linewidth]{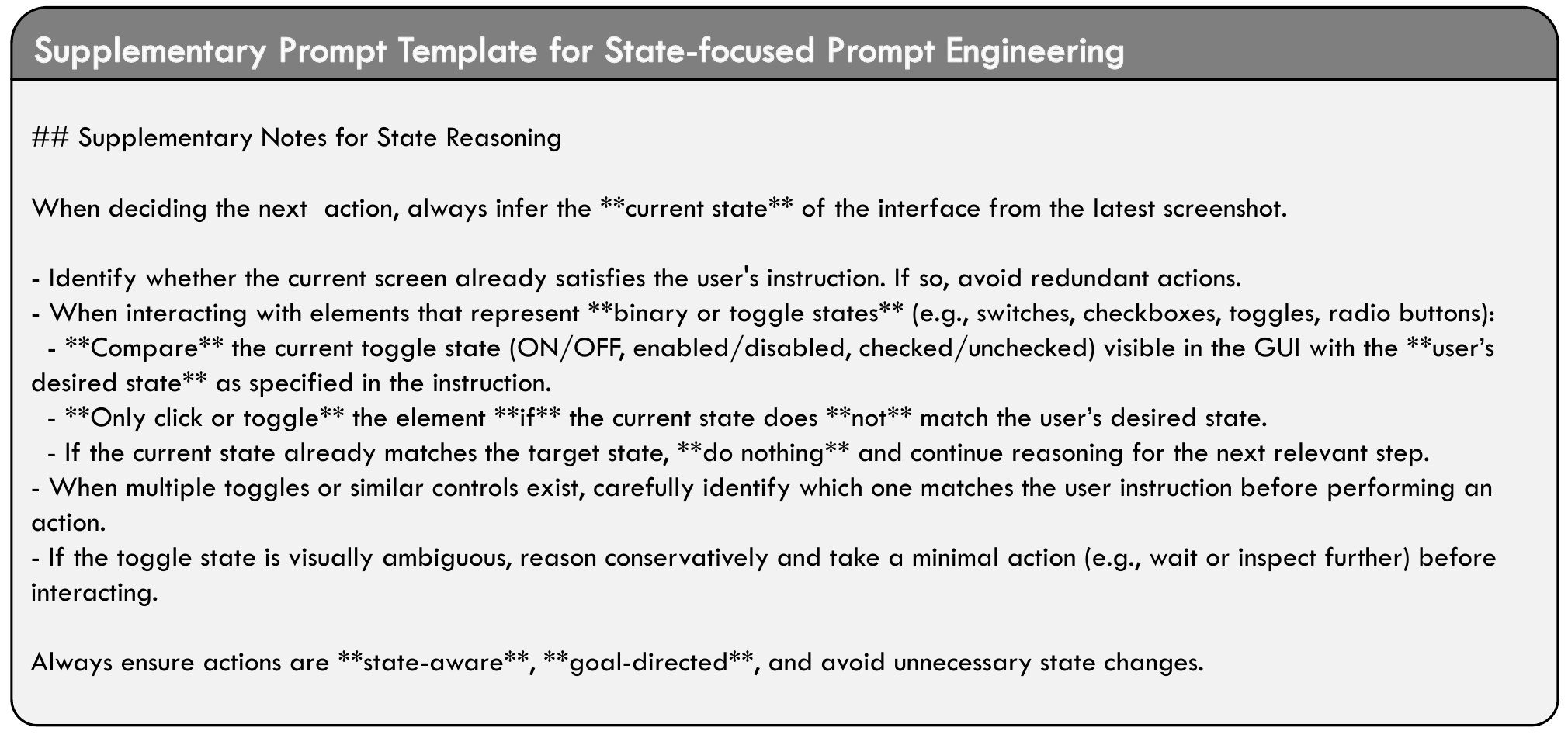}	
  \caption{Supplementary prompt template for state-focused prompt engineering.}
  \label{fig:promptStateFocusedPromptEngineering}
\end{figure*}

\begin{figure*}[!t]
  \centering
  \includegraphics[width=\linewidth]{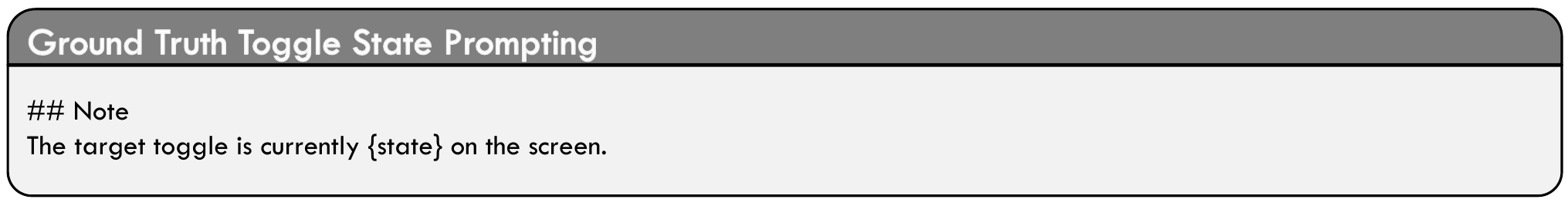}	
  \caption{Supplementary prompt template for Ground Truth Toggle State Prompting.}
  \label{fig:promptGroundTruthToggleStatePrompting}
\end{figure*}

For the two prompting baselines in Section~3.2 and Section~5.2, we append supplementary templates to the original prompts to guide the reasoning. The prompt templates for State-focused Prompt Engineering, and StaR-style Prompting are provided in Figure~\ref{fig:promptStateFocusedPromptEngineering} and Figure~\ref{fig:promptStaRStylePrompting}, respectively. For ground truth toggle state prompting in Appendix~\ref{subappendix:detailedEvaluationResultsOfExistingMultimodalAgentsOnStateControlBenchmark}, we also append supplementary templates to the original prompts to provide current toggle state and guide the reasoning. The prompt templates for ground truth toggle state prompting are provided in Figure~\ref{fig:promptGroundTruthToggleStatePrompting}.

\begin{figure*}[!t]
  \centering
  \includegraphics[width=\linewidth]{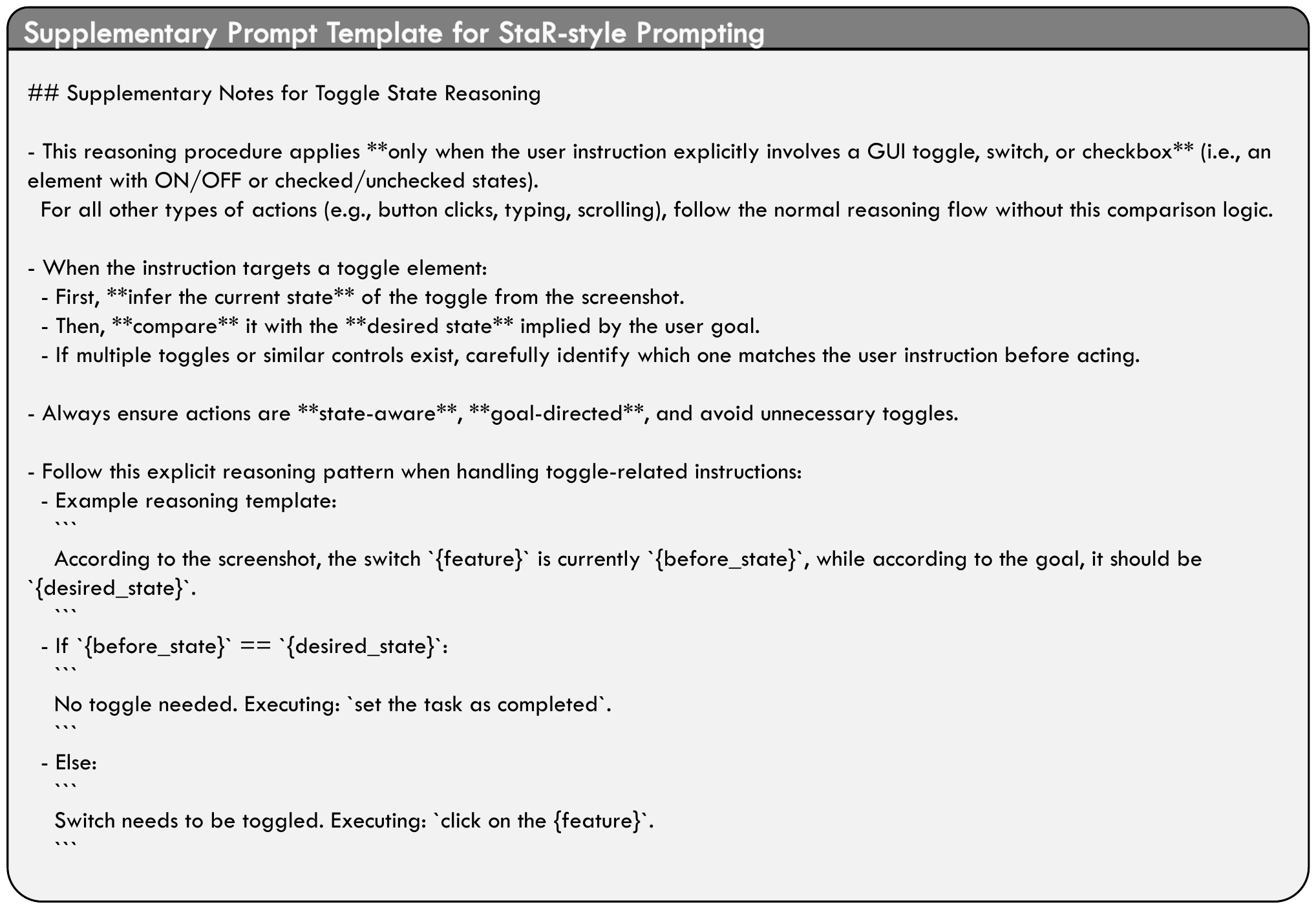}	
  \caption{Supplementary prompt template for StaR-style prompting.}
  \label{fig:promptStaRStylePrompting}
\end{figure*}

